\documentclass{article}
\usepackage{jfrExamplee}
\usepackage{amsmath,amsfonts}

\usepackage{url}
\usepackage{graphicx}
\usepackage{cite}
\usepackage{subcaption}
\usepackage{stackengine}

\usepackage[dvipsnames,svgnames,x11names]{xcolor}
\usepackage[marginparwidth=2.5cm]{geometry}

\usepackage[final]{changes} 

 \usepackage{changes} 

\title{\added[id=TF]{Optimising robotic operation speed with edge computing over 5G networks: Insights from selective harvesting robots}}

\author{
Usman A. Zahidi$^{\dagger}$\thanks{Lincoln Institute of Agri-food Technology, University of Lincoln, Riseholme Park, Lincoln, UK}\\
\texttt{uzahidi@lincoln.ac.uk} \\
\And
Arshad Khan$^{\ddagger}$\thanks{Authors contributed equally to this paper.}\\
\texttt{mukhan@lincoln.ac.uk} \\
\AND
Tsvetan Zhivkov$^*$\\
\texttt{tzhivkov@lincoln.ac.uk} \\
\And
Johann Dichtl$^*$\\
\texttt{jdichtl@lincoln.ac.uk} \\
\And
Dom Li$^*$\\
\texttt{doli@lincoln.ac.uk} \\
\And
Soran Parsa$^\ddagger$\\
\texttt{sparsa@lincoln.ac.uk} \\
\And
Marc Hanheide \thanks{School of Computer Science, University of Lincoln, Brayford Way, Lincoln, UK.}\\
\texttt{mhanheide@lincoln.ac.uk} \\
\And
Grzegorz Cielniak$^*$\\
\texttt{gcielniak@lincoln.ac.uk} \\
\And
Elizabeth I. Sklar$^*$\\
\texttt{esklar@lincoln.ac.uk} \\
\And
Simon Pearson$^*$\\
\texttt{spearson@lincoln.ac.uk} \\
\And
Amir Ghalamzan E.$^{\ddagger}$\thanks{To whom correspondence should be addressed.}\thanks{School of Computer Science and Electronic Engineering, University of Surrey, Guildford, UK.} \\
\texttt{a.ghalamzan@surrey.ac.uk} \\
\\}
\definechangesauthor[color=Red]{TF} 
\setcommentmarkup{\todo[color={authorcolor!20},size=\scriptsize]{#3: #1}}
 \newcommand{\note}[2][]{}

\begin{document}

\maketitle
\begin{abstract}

Selective harvesting by autonomous robots will be a critical enabling technology for future farming. Increases in inflation and shortages of skilled labour are driving factors that can help encourage
user acceptability of robotic harvesting. For example, robotic strawberry harvesting requires real-time high-precision fruit localisation, 3D mapping and path planning for 3-D cluster manipulation. Whilst industry and academia have developed multiple strawberry harvesting robots, none have yet achieved human-cost parity. Achieving this goal requires increased picking speed (perception, control and movement), accuracy and the development of low-cost robotic system designs. We propose the \emph{edge-server over 5G for Selective Harvesting} (\textbf{E5SH}) system, which is an integration of high bandwidth and low latency \emph{Fifth Generation} (\textbf{5G}) mobile network into a crop harvesting robotic platform,
which we view as an enabler for future robotic harvesting systems. 
We also consider processing scale and speed in conjunction with system environmental and energy costs.
A system architecture is presented and evaluated with support from quantitative results from a series of experiments that compare the performance of the system in response to different architecture choices, including image segmentation models, network infrastructure (5G vs WiFi) and messaging protocols such as \emph{Message Queuing Telemetry Transport} (\textbf{MQTT}) and \emph{Transport Control Protocol Robot Operating System} (\textbf{TCPROS}). Our results demonstrate that the E5SH system delivers step-change peak processing performance speedup of above 18-fold than a stand-alone embedded computing Nvidia Jetson Xavier NX (\textbf{NJXN}) system.

\end{abstract}


\section{Introduction}

Robotic systems are seen as a significant opportunity to help secure global food security. They can be deployed to drive environmental sustainability~\cite{pearson2022robotics} as well as economic productivity~\cite{duckett:agri-robots, marinoudi2019robotics}. Whilst many agricultural jobs have now been automated, the industry is still reliant on significant numbers of skilled human workers to hand-harvest fruits and vegetables. These skilled workers perform complex cognitive and high-fidelity manipulation tasks that have so far been evaded by robotisation~\cite{marinoudi2021future}. Developing robotic systems to automate these tasks is an urgent need in many key production regions around the world (e.g. UK, US, NL, ES, JP), which are now facing severe labour shortages \cite{Bloomberg2022, Eatingwell2022, EastAsiaForum2022}. 

These shortages are driven by the demographics of age, politics of migration, and challenging socioeconomic working conditions. Whilst robotic systems are being developed to selectively harvest essential fruit and vegetable crops (e.g., strawberries~\cite{HarvestCroo2022}, apples \added[id=TF]{ \cite{Zhang2023} and~\cite{Xiong2022} }, tomatoes~\cite{Metomotion2022}, mushrooms~\cite{Mohanan2021} and broccoli~\cite{Blok2016}), we are unaware of any commercial robot that can harvest these products at the same operational cost (or speed) as a human. \added[id=TF]{Multi-arm robots are promising alternatives for maximizing harvesting output and arms cooperation. \cite{Li2023} formulates the problem of arms cooperation and maximizing harvest through multi-arm robots as the Markov Decision Process to learn a multi-agent through reinforcement learning. \cite{Mann2016} approaches the maximum robotic harvesting problem and formulates it as an example of a k-colorable sub-graph problem for the multi-3DoF harvesting problem.}  

Key technical challenges include designing low-cost robot systems with high-speed data processing capacity. Real-time (or near real-time) data processing capacity is a significant challenge for robotic systems operating within complex and biologically diverse environments. In addition, computing must have a low cost in terms of both robot energy use and possibly carbon consumption. Here, we explore how 5G and edge computing can advance processing speed and reduce the cost of selective harvesting robotic technologies.

Recent developments in robotics, computer vision, machine learning, and communication networks enable innovation in crop-harvesting robotic platforms. For strawberries alone, multiple platforms have been developed, e.g., Dogtooth \cite{LuisaChesshire2022}, SagaRobotics \cite{Ge2020}, Harvest CROO \cite{HarvestCroo2022}, AgroRobot \cite{Agrobotics2022} and Octinion \cite{Octinion2018}. Metomotion \cite{Metomotion2022} develops greenhouse robots with a focus on tomato harvesting. Similarly, \added[id=TF]{the Certhon robots powered by Denso corporation's technology \cite{Certhon2024} and Four Growers \cite{FourGrowers}}  also develop tomato harvesting robots. Several harvesting robot projects within the research community are also proposed for various environments and other types of fruit. Robotic and Autonomous Systems for orchard fruit picking are also presented in the literature, e.g., apple~\cite{Zhang2023}, orange~\cite{Mehta2022} and \cite{Mehta2014}, grape~\cite{Luo2018}, kiwi fruit~\cite{Barnett2020} and cherry~\cite{Suraj2017}. Application of robotic harvesting for vegetables such as mushrooms and broccoli are mentioned in \cite{Mohanan2021} and  \cite{Blok2016}.

This paper presents an integration of our novel strawberry-picking robotic system~\cite{parsa2023autonomous} with a private 5G network and an edge-server. Our integrated edge-server and robotic strawberry picking system communicate over our private 5G-SA\footnote{Stand-Alone} network to facilitate near-real-time strawberry picking. This enables our strawberry-picking robot to perform at a speed comparable to a human picker. We investigated available perception approaches to find the precision closest to humans. We implemented semantic segmentation models in computationally intensive Mask-RCNN and embedded computing-friendly FBNets. These models are then deployed on both the edge-server and NJXN board. We present a comparative study on the performance of the proposed system and the NJXN board. We also present a comprehensive analysis of performance gain by quantifying the latency caused by processing, network, and model prediction. 
 
The contributions of this paper include: (1) we integrated a private 5G-SA network and an edge-server into a robotic system to process computer vision components and facilitate perception, enhancing tasks such as object detection, semantic segmentation, obstacle map, accurate 3D localisation, and mapping in near real-time. (2) We have demonstrated the effectiveness of our proposed approach in the use case of selective harvesting of strawberries. Our results demonstrate that the proposed system enables effective picking actions by a robotic system and illustrates opportunities for deploying commercial robots in fields.
We measured system deployment in a computationally intensive task involving a 3D representation of the environment to detect obstacles and enable efficient harvesting. 

We present a comparative performance analysis of MQTT and TCPROS protocols-based communication between the robots and the edge-server over WiFi and a private 5G-SA network. We also compared energy consumption and carbon emissions caused by our proposed setup and standalone embedded devices. 

Our prior work~\cite{parsa2023autonomous} has shown that processing sensory information (namely perception process) with the computing machines on our selective harvesting robot is one of the challenging improvements necessary and makes the picking process very slow. This perception process may be executed more than once for each fruit-picking cycle. Computing the perception with our robot computer (i.e., Intel Core i7 \texttrademark  CPU, 16 GB RAM, and runs Ubuntu 20.04.4 LTS (Focal Fossa))~\cite{parsa2023autonomous}, takes between 1.5 seconds to 3 seconds. The average picking time was 25 seconds. The robot takes 5 seconds to move from point home configuration to picking fruit and 5 seconds to put the fruit picked into a punnet. Motion planning can take between 20 and 1000 milliseconds. Considering only three cycles of perception (i.e., generating an action plan for the robot from sensory data) needed for completing a picking cycle, resulting in a 12-second longer picking process,  our proposed E5SH can reduce this to a maximum of 3.1 seconds. This means the E5SH robot is 5 seconds slower than our 11 seconds, our target human picking speed.

In Section~\ref{sec_review}, we provide a relevant literature review whilst the core system architecture of the proposed setup is presented in Section~\ref{sec_sys_arch}. Section~\ref{sec_exp_setup} contains details of the experimental setup followed by results and experimental evaluation presented in Section~\ref{sec_results_eval}. The paper is concluded in Section~\ref{sec_conclusion}, which also discusses the results. 

\section{Literature Review}\label{sec_review}

A comprehensive review of selective harvesting robotic systems (including hardware, perception, motion planning and motion control) is presented in~\cite{rajendran2023towards}. However, that work does not discuss the telecommunication aspect of selective harvesting robots. This section presents related works to the technical elements of our integrated system (namely telecommunications infrastructure, edge computing, computer vision and multi-process messaging protocols of E5SH).

 Processing high-dimension visual data, i.e., more than two megapixels RGB and Point Cloud (\textbf{PC}) in near-real-time presents a very challenging barrier for robotic applications where the robot needs near real-time processing for smooth performance. These processes include strawberry detection/segmentation, quality/weight/size estimation, localisation, obstacle map generation, motion planning \cite{Xiong2018},\cite{agronomy2022}, \cite{Ge2019b},\cite{Ge2020}. 
 A human expert in strawberry picking can pick 70 Kg/h and place them into a big tray, whereas an average performance is 40 Kg/h \cite{Xiong2018}. Assuming an average strawberry weighs 30 g, this means 1,333 pick actions per hour with two hands, translating into 5.5 seconds per strawberry. This means any millisecond and latency matter for data processing in technology developments. The biological latency observed in humans is 70 ms~\cite{cole1988grip}. In some cases, the robot may need to process the images of a single strawberry cluster multiple times from different views to deal with occlusion and conduct cluster manipulation for successful picking actions. Existing technologies may need between 10 seconds to more than 1 minute to pick a strawberry, making them commercially unviable.

Speed, precision, and reliability are the primary technical challenges that prevent commercialising most robotic fruit-picking systems. Fast and precise detection, segmentation, and localisation of ripe fruit in dense clusters are crucial to building successful robotic picking technologies. On the other hand, robot battery capacity and energy consumption are two factors limiting their performance. Robot energy use increases for complex tasks like strawberry picking~\cite{Agrobotics2022} as it can involve processing visual sensors data multiple times: e.g., it detects/segments and localises ripe strawberries~\cite{Ge2019b,Ge2020, tafuro2022strawberry}, builds an obstacle map~\cite{hornung2013octomap} for motion planning and plans corresponding robot actions for an effective picking movement~\cite{Motionplan2022}. This paper uses the terms `\emph{strawberry picking perception}' or `\emph{perception}' for these processes.~\added[id=TF]{Although novel learning from demonstration methods aim at reducing the computation time by directly mapping the sensor data to robot movements~\cite{tafuro2022dpmp, sanni2022deep}, such approaches are not yet mature enough to be deployed on our selective harvesting robot.} Compared to human workers, limited robot speed and precision are two significant factors of existing robotic harvesting systems that need to operate in unstructured strawberry-picking workspaces. 

~\added[id=TF]{Effective tactile sensing is essential for achieving precise and dexterous manipulation in robotics. A recent review of advanced tactile sensors tailored for selective harvesting robots is presented in~\cite{mandil2023tactile}. Notably, a novel acoustic-based tactile sensor has been developed specifically for this application~\cite{parsons2023acoustic, rajendran2024acoustic,rajendran2024ast} demonstrated for strawberry handling~\cite{rajendran202enabling}. This sensor incorporates a deformable membrane with acoustic channels, allowing it to conform to various shapes and surfaces and provide tactile feedback~\cite{vishnu2023acoustic}. It has demonstrated effectiveness in delicate pick-and-place tasks, such as harvesting strawberries. Moreover, a 2-D version of the sensor has exhibited significantly improved precision in both force localisation and normal force readings compared to prior designs~\cite{rajendran2024ast}. Novel data-driven approaches \cite{nazari2023deep} enable a selective harvesting robot to push and manipulate clusters of strawberries using tactile sensors, thereby facilitating the autonomous picking of a ripe target strawberry. In this paper, we consider only the speed challenges faced by selective harvesting robots. Our proposed solution involves an integrated edge computing system deployed over a private 5G-SA network, enabling robotic harvesting systems to operate at speeds closer to those of human pickers.}

We have also tested different perception approaches to find the best existing model in terms of precision for strawberry picking. This includes localising fruits, finding their periphery, and labelling them, which are challenging in semantic segmentation. We explored instance segmentation techniques that output pixel-level classification for targets~\cite{Ge2019a,Ge2019b}, and bounding box labelling~\cite{Onishi2019, Williams2020}. Three-dimensional localisation is also available through multi-modal comparative analysis based on YOLOv4~\cite{Ge2022}. We also explored Region-based Convolutional Neural Networks (RCNN). RCNN employs many Residual Networks (ResNets) and Region Proposal Networks for object localisation. Mask-RCNN~\cite{mask_rcnn2017}, a variant of RCNN, predicts segmentation masks along with the bounding boxes. The Mask R-CNN with 2D bounding box is inefficient for detecting and segmenting targets in real-time for strawberry picking~\cite{Ge2020}. 3D bounding box segmentation for strawberry harvesting is also unsuitable for real-time strawberry picking~\cite{Xiong2020}. The accuracy of ResNet-based segmentation is higher than other counterparts such as YOLOv3 and YOLOv5\cite{Uijlings2013}. However, they are computationally expensive, limiting the segmented images' prediction frame rate. A lighter version called~\textit{D2Go}~\cite{FBNet32021}, based on FBNetV3, is proposed using a simplistic Differential Neural Architectural Search. It also renders device-aware training and quantisation for mobile devices. Although {D2Go} has faster inference times, it has lower accuracy than Mask-RCNN~\cite{HastyAI2022}. 

Image segmentation and building a 3D occupancy grid mapping are required for planning and picking actions for the robotic harvesting system. We use the OctoMap library for this purpose, which is a part of Robot Operating System (ROS)~\cite{quigley2009ros} packages \footnote{\url{https://wiki.ros.org/octomap}}. The OctoMap library provides data structures and mapping algorithms required to model arbitrary environments only based on sensory information. The representation models occupied areas as well as free space. The distinction between free and occupied space is essential for safe robot operation~\cite{hornung13auro}.

Whilst 5G communication in agriculture shows theoretical promise, due to its emerging nature and capital cost, there have been relatively few published use cases~\cite{van20225g,zhivkov:5g:ukras,zhivkov-et-al-machines:2023}. 5G, like other telecommunication networks, is set up using a particular-sized cell. Each cell can connect to many devices where broadband sharing/slicing is built into the core technology. We used a private 5G-SA N77 system with a small (micro) cell, which has a medium range reach (c. 200m-1km) with bandwidths of $>$50 Mbps and latency of $<$20ms~\cite{zhivkov-et-al-machines:2023}. 

We evaluated the performance of two different messaging protocols: i. TCPROS and ii. MQTT. TCPROS is the transport layer used by ROS, based on the standard TCP protocol \cite{cerf:tcp} for communicating messages. ROS communication is based on the publish-subscribe messaging pattern \cite{quigley:ros}. TCPROS differs from standard TCP as it has a specific connection header definition used to identify ROS-specific communication parameters, such as subscriber name, topic, and message definition. These necessary fields are defined in the TCPROS header information. TCPROS is similar to TCP; hence, it does not add overhead or complexity to data transmission. Nonetheless, there are known security flaws within TCPROS, such as weak identity verification between communicating devices with the transmission in plain text, as highlighted in \cite{beibei:ros-security}. 
Another publish-subscribe messaging passing protocol is MQTT. MQTT is a standard messaging protocol with the latest version, MQTT 5.0 \cite{mqtt:oasis}, sitting on top of the TCPROS transmission protocol. MQTT uses TCPROS to establish a broker. The broker redirects messages from the publisher to the correct subscriber. However, unlike TCPROS, communications can be secured from publisher to subscriber. MQTT is designed for the guaranteed receiving of messages by subscribers and publishers. Hence, we use MQTT-based messaging named QOS0 and QOS1. 

Edge computing refers to using a server relatively close to where data collection is performed (for data processing and/or storage). The goal is to reduce latency and increase reliability for tasks that cannot be performed on-site (e.g., on the robot). 
For instance, Chen et al. \cite{chen2021fogros} introduced FogROS, a framework to ease the deployment of tasks (e.g. SLAM computations) on a cloud server. 
Antevski et al.~\cite{antevski2018enhancing} used edge computing for analysing WiFi, quality, connectivity, and reliability.  
Hayat et al. \cite{hayat2021edge} used a 5G network to offload the computation to an edge-server necessary for navigating drones. The study showed there are cases where edge computing is significantly beneficial. Huang et al. \cite{huang2022edge} used edge computing in the context of multi-robot collaborative SLAM, called RecSLAM, a 2D SLAM algorithm that works on the edge-server, outperforming state-of-the-art solutions. However, these works do not deal with robotic manipulation involving cases with physical robot interactions. These cases may need multiple queries for the perception module to segment/identify and localise a ripe fruit.  

\section{System Architecture}\label{sec_sys_arch}

\begin{figure*}[tb!]
\centering
\includegraphics[width=\linewidth]{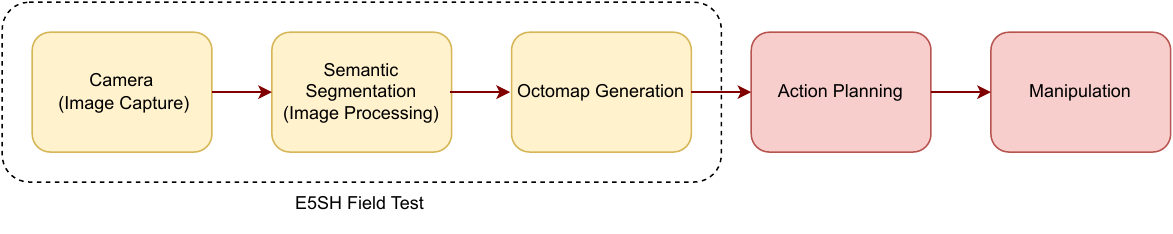}
\caption{\textbf{System process pipeline:} An overview of the E5SH system process pipeline overview, processes in yellow are field-tested during E5SH experiments. The processes shown in red were not completed in field testing due to strawberries' out-of-season time.}
\label{fig_fastpick_process_pipeline}
\end{figure*}

The processing pipeline for a robotic harvesting system is shown in Fig. \ref{fig_fastpick_process_pipeline}. This pipeline includes image acquisition, semantic segmentation, OctoMap generation, action planning and manipulation execution. Our prior works~\cite{parsa2023autonomous, tafuro2022strawberry} indicated the challenges in the first three blocks of robot perception to robot action pipeline\footnote{ Because the time we tested the E5SH system was off-season, the action planning and manipulation were not performed during the E5SH field tests.} (shown in Fig.~\ref{fig_fastpick_process_pipeline}). Hence, we only field-tested the three first left blocks in~Fig.~\ref{fig_fastpick_process_pipeline} for estimating the overall speedup. The overall performance of E5SH is estimated by aggregating the results of our field tests and action planning and manipulation statistics from the references above.

\begin{figure*}[tb!]
\centering
\includegraphics[width=\linewidth]{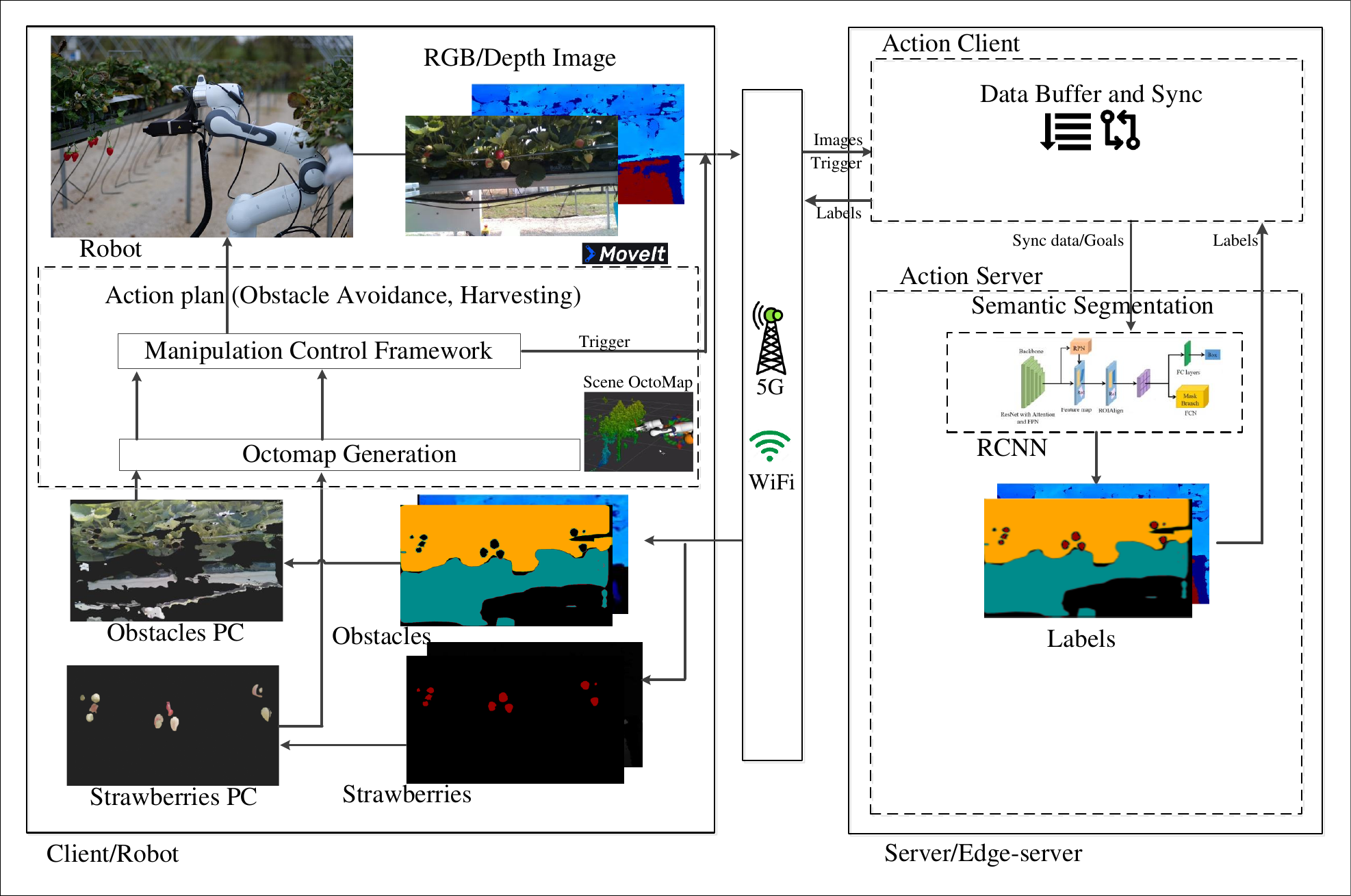}
\caption{\textbf{Detailed system architecture of the proposed setup :} Illustration of the data flow between robot and edge-server over 5G and WiFi. The edge-server side performs 4-class classification. The computing device on the robot performs localisation, OctoMap generation and action planning. The robot performs the picking action, collects the images, and sends them to the edge-server. }
\label{fig_fastpick_flow}
\end{figure*}

The system architecture of the demonstration setup is shown in Fig.~\ref{fig_fastpick_flow}. Our system comprises a robot platform, an arm with a gripper, communication devices, a private 5G-SA system, an edge-server, and a robot-mounted laptop for the OctoMap generation. We allocated heavy computing tasks, such as four-class segmentation, to the edge server. Data transfer from the robot to the edge-server is limited to the camera stream (RGB/Depth images and camera info topics) and service call (/trigger) to indicate when the robot needs an update of the segmented depth images. On the server, a ROS action client receives the continuous stream of images and synchronises that data. Once a service call is received, the action client sends an action goal containing the synchronised camera data to the action server. The action server runs the Detectron2 (Mask-RCNN) and Detectron2Go (D2Go) models and segments the depth images to four labels, i.e. ripe strawberry, rigid obstacle, soft obstacle (canopy), and background. The result is returned to the action client, which publishes the labelled depth images. The depth images are returned to the robot that constructs 3D point clouds of respective obstacles and strawberry segments. These point clouds are then employed to generate corresponding OctoMap (Fig.~\ref{fig_fastpick_flow}).

\noindent Next, the OctoMap of the obstacle are fed into the navigation control framework, which performs action planning of the robot. Similarly, strawberry OctoMap is passed to the manipulation control framework. This ensures that harvesting is performed with accurate localisation so that neighbouring strawberries are not considered obstacles during operation. The whole pipeline is triggered by the trajectory planner on the robot, which relies on the labelled depth images for obstacle detection.

\subsection{Semantic Segmentation} 
\label{sec:sem-seg}
\begin{figure}[tb!]
\centering
\begin{subfigure}[t]{0.4\textwidth}
\includegraphics[width=\textwidth]{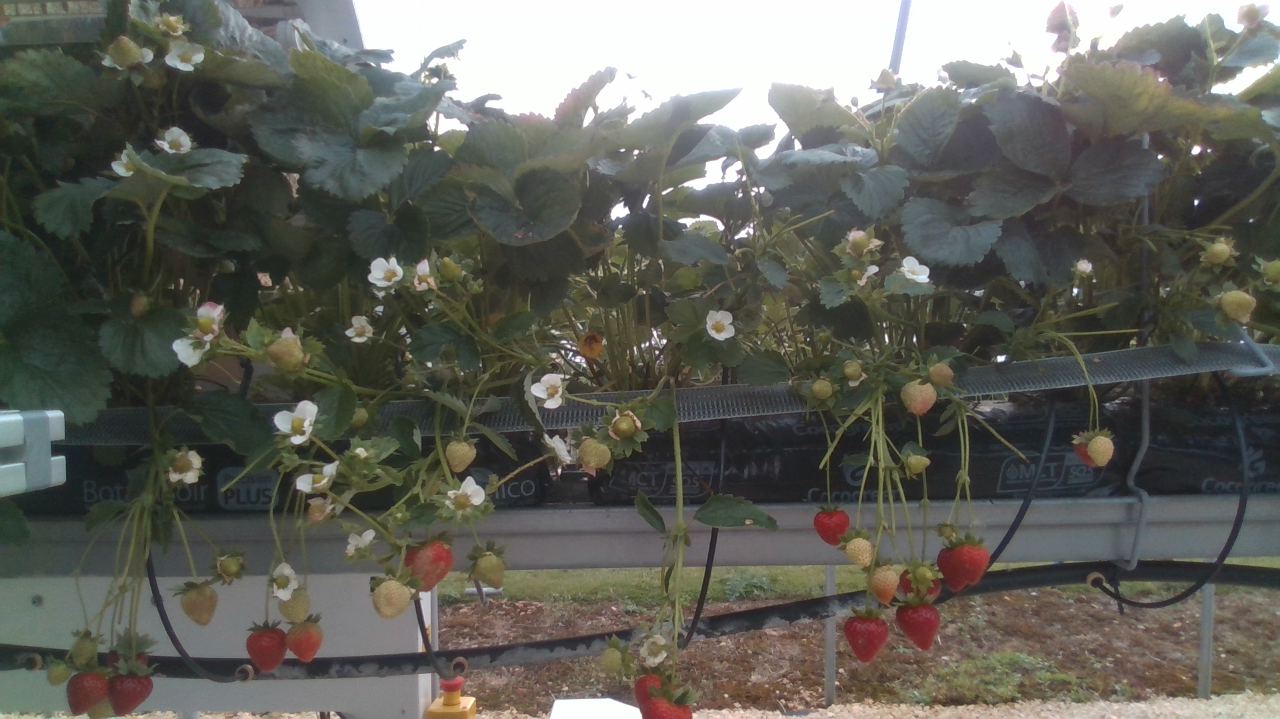}
\caption{}
\hfill
\end{subfigure}
\begin{subfigure}[t]{0.4\textwidth}
\includegraphics[width=\textwidth]{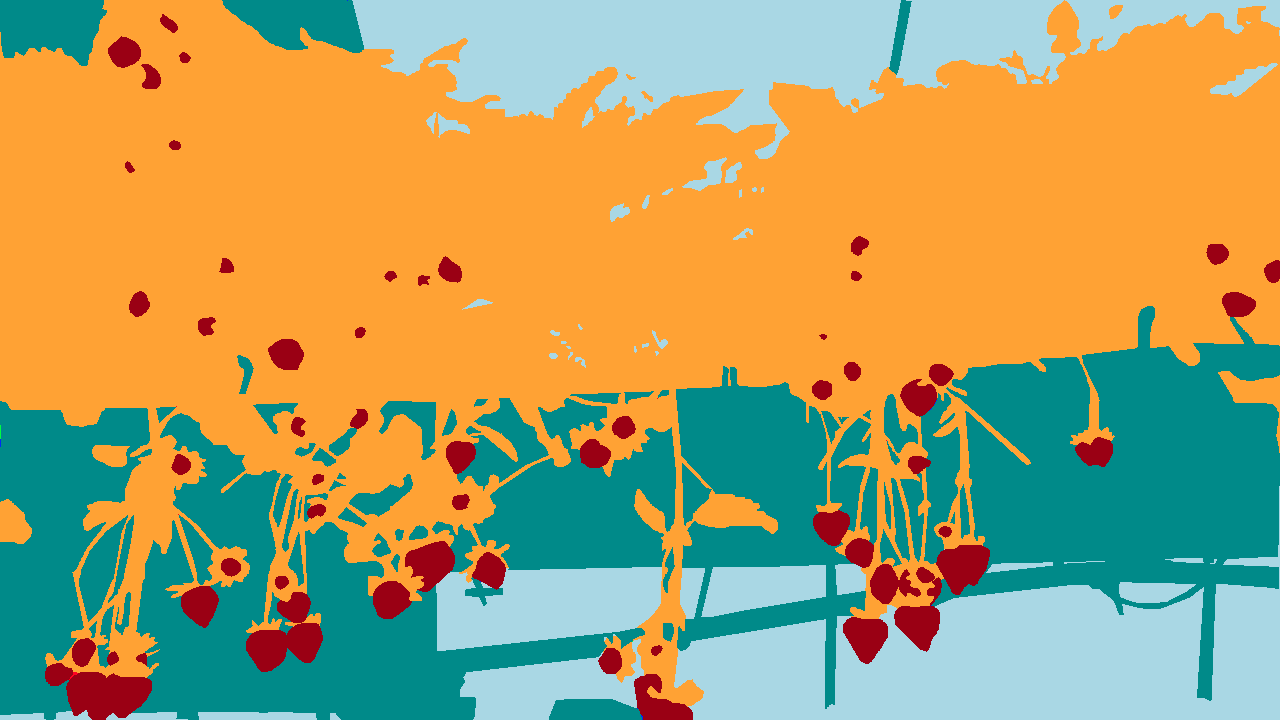}
\caption{}
\end{subfigure}

\begin{subfigure}[t]{0.4\textwidth}
\includegraphics[width=\textwidth]{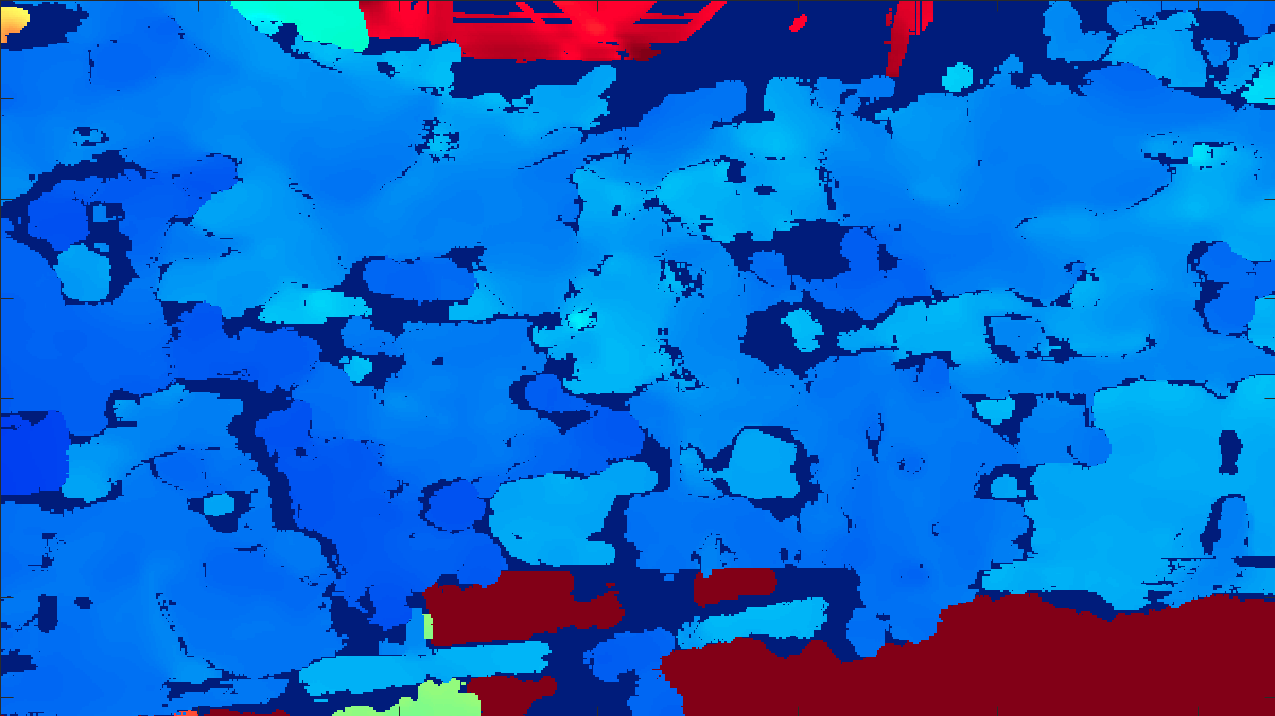}
\caption{}
\hfill
\end{subfigure}
\begin{subfigure}[t]{0.4\textwidth}
\includegraphics[width=\textwidth]{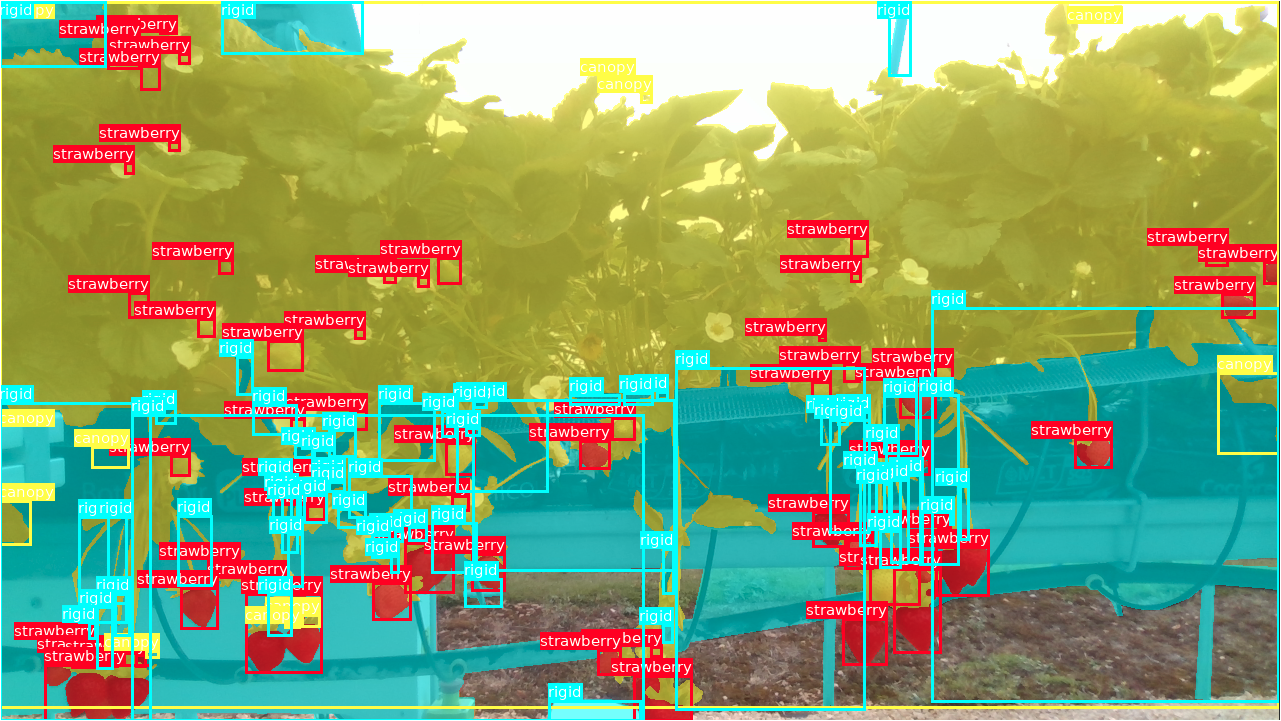}
\caption{}
\end{subfigure}
\caption{{E5SH Image annotation: (a) Original image; (b); semantic annotation at the later season (September) of strawberry under direct/passive solar illumination; (c) depth image; (d) equivalent bounding box annotation.}}
\label{fig_straw_labelled}
\end{figure}

Robots in our setup traverse the polytunnel where strawberries are grown, and images are acquired from July to October 2021. A sample image and its corresponding ground-truth (GT) annotation, the depth image, and bounding box annotations are shown in Fig.~\ref{fig_straw_labelled}. We trained Mask-RCNN and FBNetv3 by employing Detectron2 and D2Go and measured prediction, accuracy, and latency between edge and embedded platforms. Moreover, we measured the performance of our system with a bespoke, data-intensive perception task by deploying semantic segmentation to the edge-server for crop-harvesting robotic applications. These images are segmented according to the list of classes: (1) \textit{Strawberries}:	Both ripe and unripe strawberries are included in this class, while flowers and nascent strawberries are not included; (2) \textit{Canopy}: The strawberry plant is labelled as the Canopy; (3) \textit{Rigid Obstacle}: Any object which the robot should avoid is a rigid obstacle. It includes metal structures,	pipes, and humans; (4) \textit{Background}: The remaining region is marked as background. 

The harvesting operation requires obstacle avoidance, which is classified as hard obstacles. The canopy is a soft obstacle, which means the robot can interact with and push it to some extent. Hence, we classified it as a soft obstacle (or canopy). We need to precisely localise the ripe strawberries as our robots need to reach the stalk of the fruit and grip/cut it~\cite{parsa2023autonomous}. We also classified the ripe fruits, whereas the remaining region of the image is classified as background. The semantic segmentation model provides segmentation masks for soft objects such as fruits and leaves and hard objects such as canopy, poles, and other background objects. Depth masks were used to create point clouds and OctoMap for segmented masks; therefore, they are time-efficient and precise and facilitate unhindered action planning. 

\added[id=TF]{Instance segmentation detects multiple instances of the same label in an image, which is one step ahead of semantic segmentation, which only labels all the pixels in an input image for a specific label. Since the target picking object was always strawberry during this study, we opted for semantic segmentation and contour detections to find the strawberries if there are multiple strawberries within an input image. The contour detection on the mask of the strawberry label also allows the removal of any strawberries that are not the target of picking strawberries from the OctoMap. Removing non-target strawberries from the OctoMap results in a high successful planning rate because we assume that non-target strawberries are soft objects that can be pushed during the pick motion action.}

\subsection{Computational Platform}
\label{sec:computation}
We compare the performance of our system when using either an onboard NJXN (i.e. on the robot) or an edge-server networked to the robot's controller, described in Section~\ref{sec_exp_setup}.
Our \textbf{edge-server} is an Intel(R) Core(TM) i7-9700K CPU @ 3.60GHz, an eight-core machine equipped with 64 GB memory and Nvidia RTX-2080 GPU having 10 GB GDDR memory. It has Ubuntu 20.04 hosting ROS noetic and Nvidia CUDA 11.1. Facebook Detectron 2.0.6 framework is installed with torch 1.10.0 and D2Go 0.0.1 for semantic segmentation. 
Our onboard processor is an NVIDIA Jetson Xavier (NJXN) platform.

\subsection{Network Infrastructure}
\label{sec:network-inf}
As outlined in Section~\ref{sec_exp_setup}, two different network infrastructures were compared, including a private 5G-SA network and standard WiFi. The configuration of each is described below.

\begin{figure}[tb!]
\begin{subfigure}[tb!]{1.0\textwidth}
    \centering
    \includegraphics[scale=0.35]{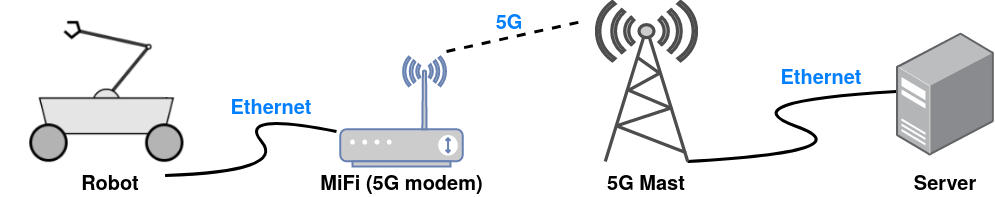}
    \caption{}
    \label{fig_network_topo_g52}
\end{subfigure}\\
\begin{subfigure}[tb!]{1.0\textwidth}
    \centering
    \vspace{1cm}
    \includegraphics[scale=0.35]{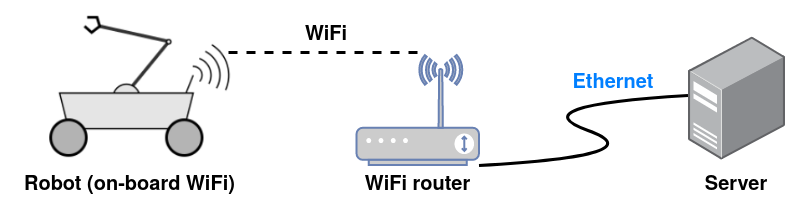}
    \caption{}
    \label{fig_network_topo_wifi}
\end{subfigure}
\caption{The basic network topology for communication from the robot to the edge-server for (a) 5G network and (b) WiFi network.}
\label{fig_network_topo_g5}
\end{figure}

\textbf{5G Network.} 
We use a private 5G-SA (N77), also called the sub-6GHz band, network system with operational parameters listed in Table~\ref{tab:5gsystem}. Since the robot has no 5G capabilities, we use a 5G-capable MiFi router. The router can be either directly attached to the robot or placed in a secure location in the field to serve multiple robots as the access point. This router then establishes a 5G connection to the mobile 5G tower, connected to a local LAN (Fig.~\ref{fig_network_topo_g5}). Communication between the two endpoints is handled via the ROS mqtt\_bridge module, which uses the MQTT protocol for machine-to-machine data transfer.

\textbf{WiFi Network.}
Similar to the 5G network topology, the WiFi topology is specially designed for robot communication in the field at the Riseholme campus, University of Lincoln. Our mobile platform (Thorvald II robot, SagaRobotics Ltd) is connected to the WiFi network to establish a connection with the edge-server (Fig.~\ref{fig_network_topo_wifi}).
However, expanding the network can be achieved by attaching a WiFi access point in a secure location in the field to serve multiple robots or a robot can be directly connected as it has WiFi capabilities onboard.
A WiFi access point must be connected to the central network for communication.
The same protocol is used as in the 5G case, the MQTT protocol for machine-to-machine data transfer.

\section{Experiments} \label{sec_exp_setup}

We conducted experiments to evaluate several different aspects of the performance of our system:
(a) comparison of semantic segmentation models (as described in Section~\ref{sec:sem-seg});
(b) comparison of NJXN vs edge-server platforms (described in Section~\ref{sec:computation}), for running the trained semantic segmentation models;
and
(c) comparison of communication network infrastructure (described in Section~\ref{sec:network-inf}) and messaging protocols (MQTT vs TCPROS).
Results of these comparative experiments are presented in Section~\ref{sec_results_eval}, following details of our experimental setup, provided below.

\begin{figure}[tb!]
    \centering
    \includegraphics[scale=0.365]{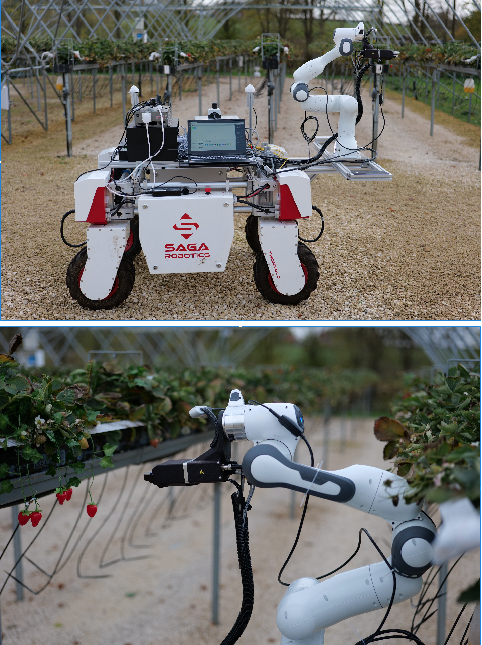}
    \caption{The robot and polytunnel facility at the University of Lincoln: A Franka Emika arm equipped with the strawberry picking end-effector \cite{parsa2023autonomous} sits on a Thorvald II. Cameras, communication devices, and laptops are also shown. }
    \label{fig_thorvald}
\end{figure}

Our experimental setup is located on the Riseholme Park farm campus, home to the Lincoln Institute for Agri-food Technology (LIAT) at the University of Lincoln, UK.
The farm includes a test facility for growing strawberries, consisting of two polytunnels. In 2022, these tunnels produced more than 2,000 kg of fruit. Each tunnel has five rows of tabletop strawberries in three different varieties, e.g., Zara, El-Santa, and Ever-bearing.  
We performed several field tests and collected the corresponding data and images across different growth stages. These images are taken from an industrial grade RGB-D \textregistered Intel D435i \texttrademark RealSense camera mounted on a Thorvald II robot as shown in Fig.~\ref{fig_thorvald}. A Franka Emika Panda arm having seven degrees of freedom (DOF) and a strawberry picking end-effector~\cite{parsa2023autonomous} are used for picking actions.

We performed approximately twelve experiments on seven different days and times under different climate conditions to ensure coverage of variance in network communication, illumination conditions, and strawberry growth stages.

\subsection{Image Data Set}

A preliminary sample size of three thousand RGB-D image sequences was acquired and registered. A sub-set of around 212 images is selected using~\cite{Zahidi2021} and annotated by third-party professional annotators~\cite{RedbrickAI2021} to create ground truth. Experiments presented in this paper are performed on $142$ training set images taken at different times of day, seasons, months, and illumination conditions in the presence of hard and soft shadows, inside polytunnels, and with several different camera orientations. The test set includes $70$ images. The dataset contains 92 \% images with 1280$\times$720 resolution, and the other 8 \% images have VGA resolution. The test dataset is utilised to evaluate the models' performance and distribution. The live network transmission throughput and prediction time delay were logged at thirty frames per second, at the frame resolution of 848$\times$480 for both RGB and depth images.

\textbf{Data set analysis.}
Once the training and test dataset candidate images were selected, the metadata, such as instance counts for each class, and the bounding box’s mask area distribution pixel ratio, was collected to detect the potential data imbalance. The distribution between rigid and strawberry classes is balanced; however, the canopy class is smaller in number. The reason for the lower number of the Canopy is its more prominent area. The pixel ratio implies a relatively uniform region occupancy between rigid obstacles, Canopy, and background. The strawberries' abundance is only within 2.2\% pixels.

\begin{table}[!t]
\centering
\scalebox{0.8}[0.8]{
\begin{tabular}{|c|c|c|c|c|}
\hline
\multicolumn{1}{|c|}{\textbf{Measurement}} & \textbf{Rigid Obstacles} & \textbf{Canopy} & \textbf{Strawberries} & \textbf{Background}       \\ \hline
\multicolumn{1}{|c|}{Pixel ratio (\%)}            & 35.9  & 33.3 & 2.2 & 28.7          \\ \hline
\multicolumn{1}{|c|}{Bounding box instances}      & 1501 & 526 &  1464 & N/A                  \\ \hline
\end{tabular}}
\caption{Measurement of \emph{per class abundance} in terms of pixels percentage and bounding box instance in complete training and test datasets. }
\label{tab:dataset_cls_instances}
\end{table}

\textbf{Training Parameters.}

We train Detectron2 using the base model of Mask RCNN R101 FPN 3x. The model is trained in 90000 iterations with a learning rate of 0.0025, and the test threshold is set to 0.5. We used a Stochastic Gradient Descent optimiser during the training. Similarly, we train D2Go using the Mask RCNN FBNet v3a C4 with the same learning rate, number of iterations, test threshold, and optimiser. The instance distribution of the dataset is given
in Table \ref{tab:dataset_cls_instances}. The distribution between rigid and strawberry classes is balanced; however, the canopy class is smaller in number. We applied several augmentation methods on training and test datasets, such as random flip, cropping and brightness, for improved training on our dataset.

\subsection{Evaluation Metrics} \label{sec_eval_metrics}

For comparing the accuracy of the trained semantic segmentation models, we employ the standard evaluation metrics for the experimental comparisons~\cite{eval_metrics_2021}. Pixel-wise True Positive ($T_P$), False Positive ($F_P$) and False Negatives($F_N$) are computed for all images. Precision ($p$), recall ($r$) and F1-measure ($F1$) for all classes are defined as
$p=\frac{T_P}{T_P+F_P}$, $r=\frac{T_P}{T_P+F_N}$ and $F1=\frac{2pr}{p+ r}$, which were calculated for each class separately. The Average Precision is $AP=\frac{1}{n} \sum_{i=1}^n p\left(\tau_i\right)$ and Average Recall is $AR=\frac{1}{n} \sum_{i=1}^n r\left(\tau_i\right)$ where $\tau$ is the threshold function. 
We present the pixel-wise results and discuss them in Sections~\ref{sec:results:sem-seg}
and~\ref{sec:results:seg-acc}.

To compare the processing speed of the different computational platforms, we consider the number of prediction \emph{frames per second} (FPS) each configuration can handle.
We focus on the \emph{speedup} obtained on the faster edge-server versus the onboard NJXN processor.
Results are presented in Section~\ref{sec:results:comp-speed} for the different semantic segmentation models running on each processor. Additional experiments were conducted to simulate each processor servicing one vs. three robots simultaneously to stress the processor capacity. For comparing the performance of the different communication networks, we measure the \emph{latency}
Results are presented in Section~\ref{sec:results:network}.

\section{Results} \label{sec_results_eval}

Our results include the speed of perception processing, cumulative delays in image capturing, transmission, semantic segmentation and reception for both our proposed \emph{edge-server over 5G selective harvesting} (E5SH) system and the NJXN board. The parameters included in our study are (1) mode of communication (5G/WiFi), (2) underlying protocols (MQTT/TCPROS), and (3) computation throughput of edge-server and embedded device during the segmentation task. As the quality of semantic segmentation is essential to the performance of 3D localisation, we present and discuss the performance measure of the segmentation models, i.e. Detectron2, D2Go 8-bit (\textbf{D2Go-8}) and D2Go 32-bit (\textbf{D2Go-32}).  

\subsection{Scene segmentation}
\label{sec:results:sem-seg}
\begin{figure*}[t]
    \includegraphics[width=\linewidth]{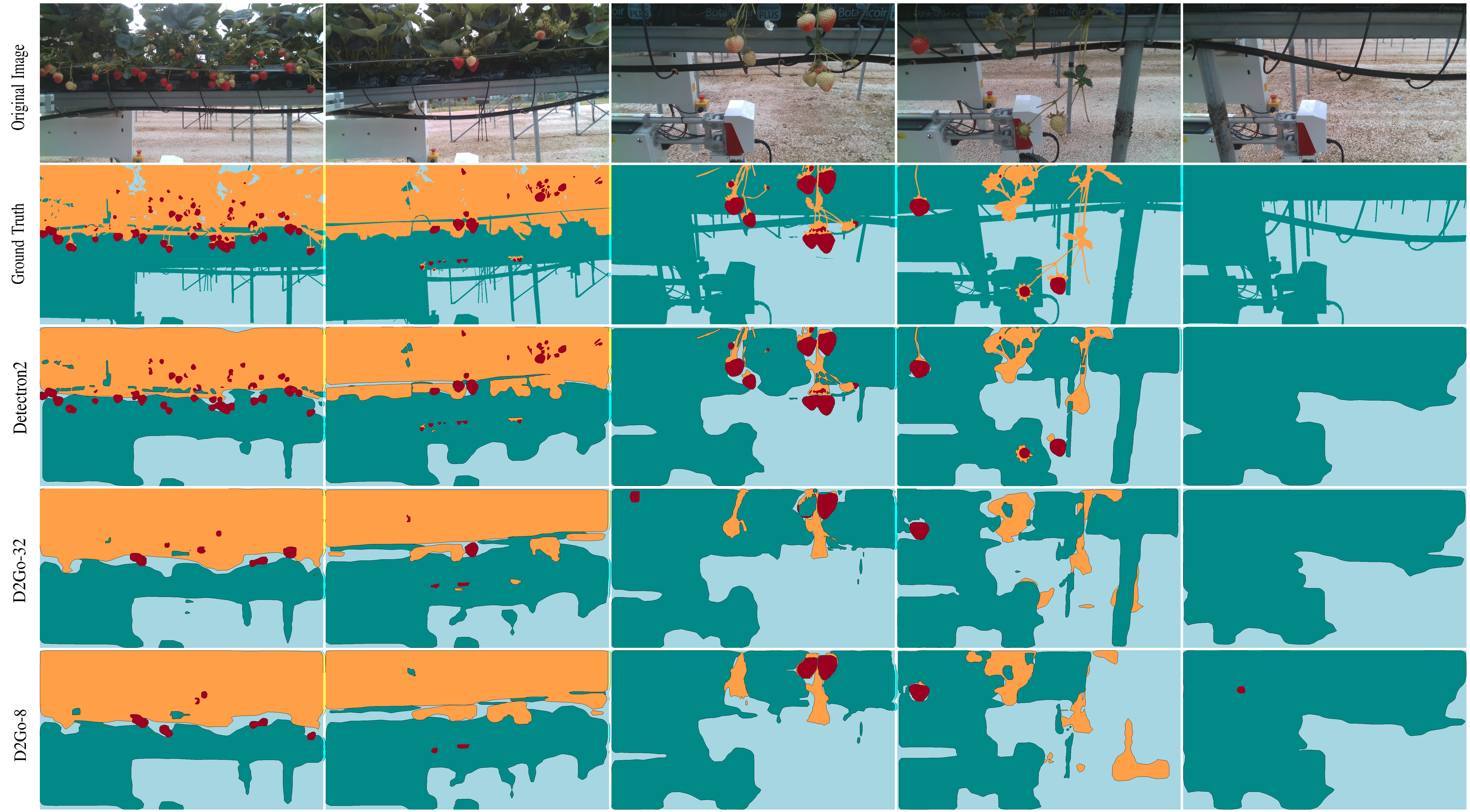}
    \caption{Each column depicts in order; Original Image, Ground Truth, Segmented output from Detectron2, D2Go-32, D2Go-8, respectively.}
    \label{fig_seg_output1}
\end{figure*}

A sample of our qualitative results is shown in Fig.~\ref{fig_seg_output1}, which shows the visual output of five images taken at different growth stages, illumination conditions, time of day and camera orientations. The segmentation output in Fig.~\ref{fig_seg_output1} shows that the standard and optimised D2Go have missed several strawberries. In contrast, Detectron2 correctly classifies most strawberries (Fig.~\ref{fig_seg_output1}). None of the segmentation models could correctly segment the sharp edges of the rigid obstacles represented by similar regions with a smooth periphery. Although the models miss the region of sharp edges in large canopy areas, Detectron2 has an improved classification for branches. Qualitatively, both Detectron2 and D2Go-32 perform well in segmenting partially visible strawberries. In contrast, the optimised D2Go-8 classifies a red cap on the robot as a strawberry.
 
\subsection{Segmentation Accuracy}
\label{sec:results:seg-acc}
\begin{figure}[tb!]
    \centering
\begin{subfigure}[tb!]{0.49\textwidth}
    \centering
    \includegraphics[width=\textwidth]{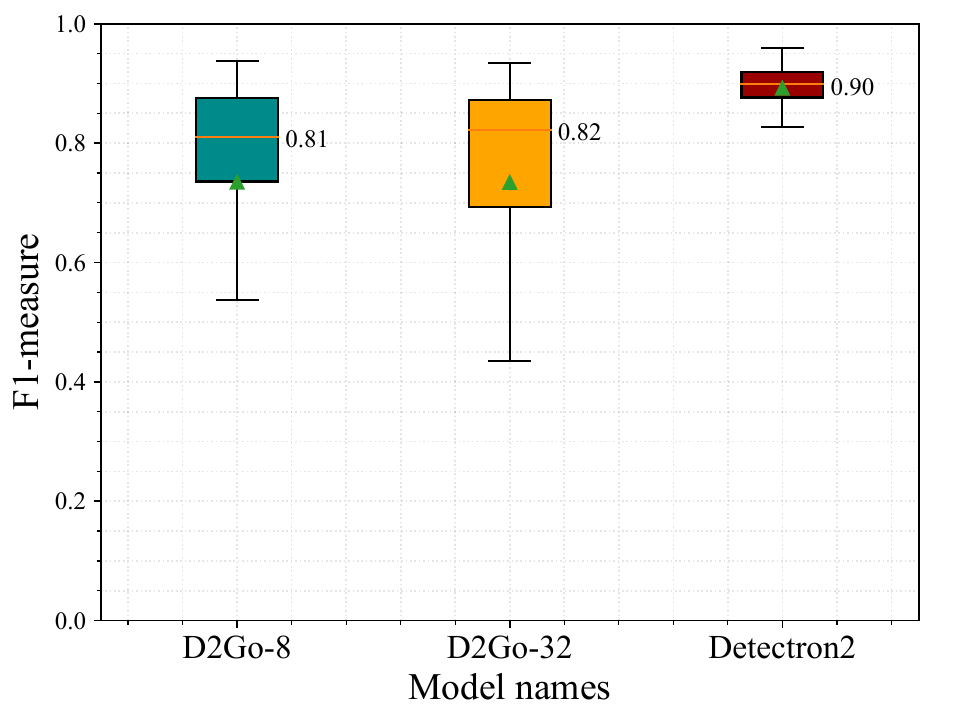}
    \caption{}
\end{subfigure}
\begin{subfigure}[tb!]{0.49\textwidth}
    \centering
    \includegraphics[width=\textwidth]{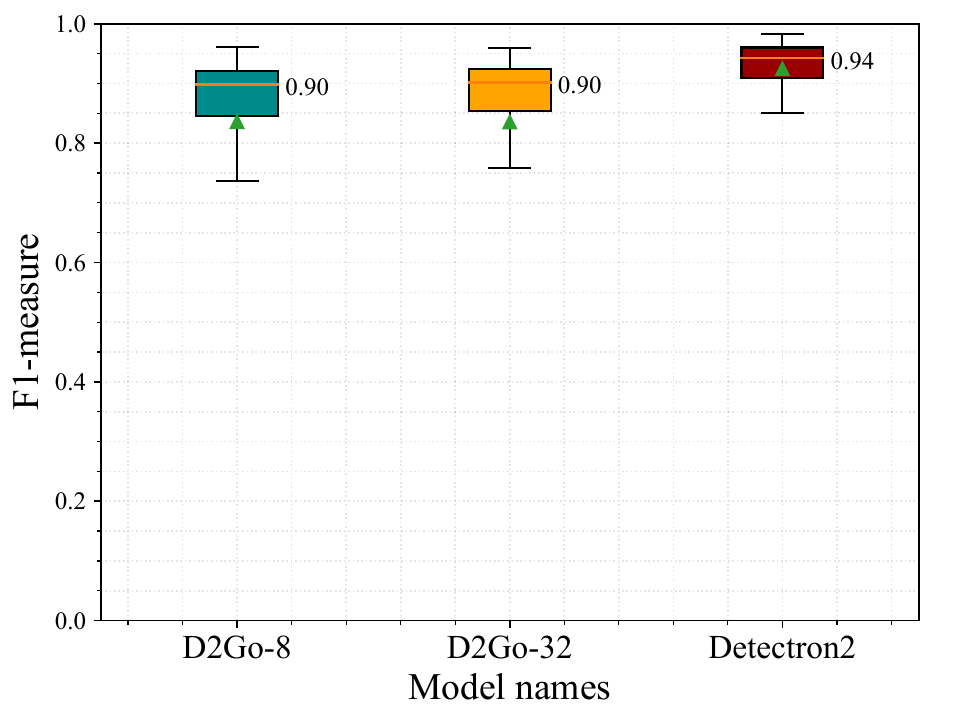}
    \caption{}
\end{subfigure}

\begin{subfigure}[tb!]{0.49\textwidth}
    \centering
    \includegraphics[width=\textwidth]{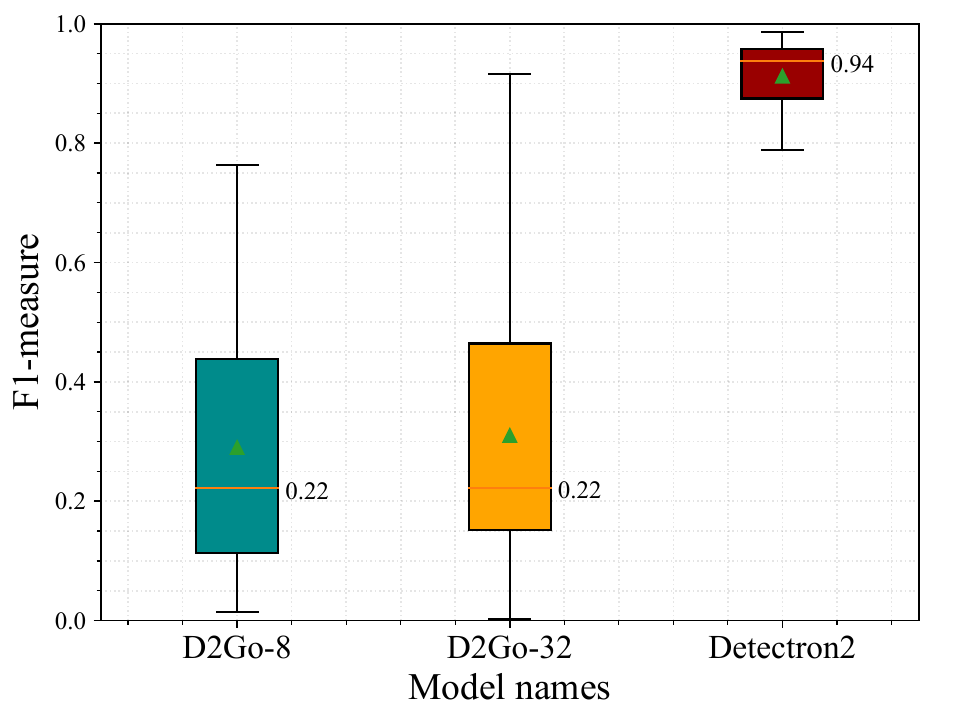}
    \caption{}
\end{subfigure}
\begin{subfigure}[tb!]{0.49\textwidth}
    \centering
    \includegraphics[width=\textwidth]{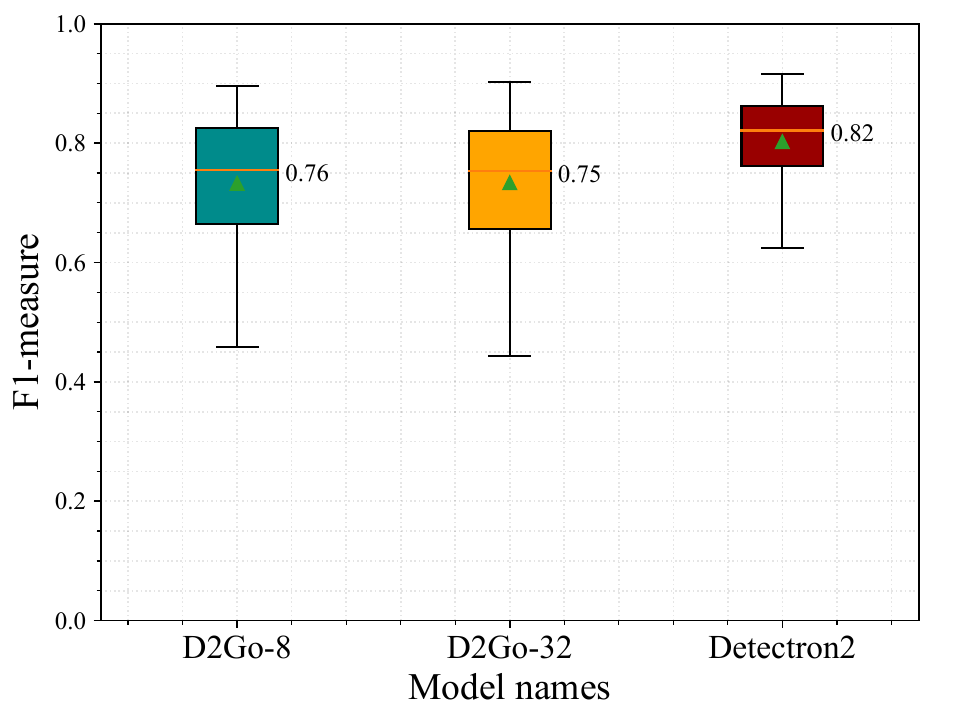}
    \caption{}
\end{subfigure}

\caption{Data distribution for F1-measure of individual segmentation models (\textbf{D2Go-8 D2Go-32 and Detectron2 models}, respectively) for (a) Strawberry (b) Canopy (c) Rigid Obstacles and (d) Background. The distribution is for the test dataset.}
\label{fig_avg_f1}
\end{figure}

Our quantitative analysis assesses segmentation quality using \emph{F1-measure} that combines precision and recall. Segmentation performance is evaluated by metrics like per class abundance,e.g., in terms of pixel percentage, discussed in Section~\ref{sec_eval_metrics}. We considered four classes for the model to predict, as mentioned in Sec.~\ref{sec:sem-seg}. 
The distribution of the F1-measure in the test dataset is given in Fig.~\ref{fig_avg_f1}. The evaluation is based on the pixel-wise calculation of true positive $T_P$, false positive $F_P$, and false negative $F_N$ concerning the Ground Truth. 
In general, both D2Go models have similar F1-measure distributions, as shown in Fig. \ref{fig_avg_f1}. In this context, the centre of distribution for \emph{rigid obstacle's} class is the lowest, with a median of 0.22 for D2Go-8 and D2Go-32, where the overall distribution is positively skewed. The Inter Quartile Range (IQR) of the rigid obstacle class for both these models is between (0.11-0.44) and (0.15-0.47), respectively. Our results show that D2Go has the poorest segmentation quality for rigid obstacle class. In contrast, the Detectron2 model has a median F1-measure of 0.94, which considerably improves over the D2Go F1-measure of 0.22. Our qualitative analysis (e.g. shown in Fig.~\ref{fig_seg_output1}) also agrees with this.  
The \emph{background} class yields the F1-measure's median of 0.76 and 0.75 for D2Go-8 and D2Go-32, respectively. They also possess similar IQRs, which lie between 0.69 and 0.81, negatively skewed. Here, Detectron2 outperforms D2Go-8 and D2Go-32 with an F1 median of 0.82 and a lower IQR of 0.76 and 0.87.

The F1-measure medians of the distribution for the \emph{\textit{canopy}} class increase to 0.90 for both D2Go models. Furthermore, their distribution remains negatively skewed, with a similar IQR between 0.85 and 0.96. On the other hand, Detectron2 again outperforms others with the F1-measure median of 0.94 and IQR within 0.91 and 0.97. 

The F1-measure distribution for \emph{strawberry} in D2Go versions is similar, with similar medians of 0.81 and 0.82 for D2Go-8 and D2Go-32, respectively. D2Go-32 has a larger IQR (0.69-0.88) than D2Go-8, which ranges between 0.78-0.87. Interestingly, Detectron2 better segments the strawberry class with a median of 0.90 and bearing lower data IQR (0.87-0.92). Detectron2 obtains a much better strawberry segmentation quality, helping with improved 3D localisation, which is very useful for harvesting robots.
Our results demonstrate that Detectron2 outperforms D2Go-8 and D2Go-32 regarding segmentation quality for obstacles and strawberries. D2Go-32 comes after Detectron2 in terms of quality performance with a slight edge over D2Go-8.

\subsection{Computation Speed}
\label{sec:results:comp-speed}
\begin{figure}[tb!]
\centering
\includegraphics[scale=0.5]{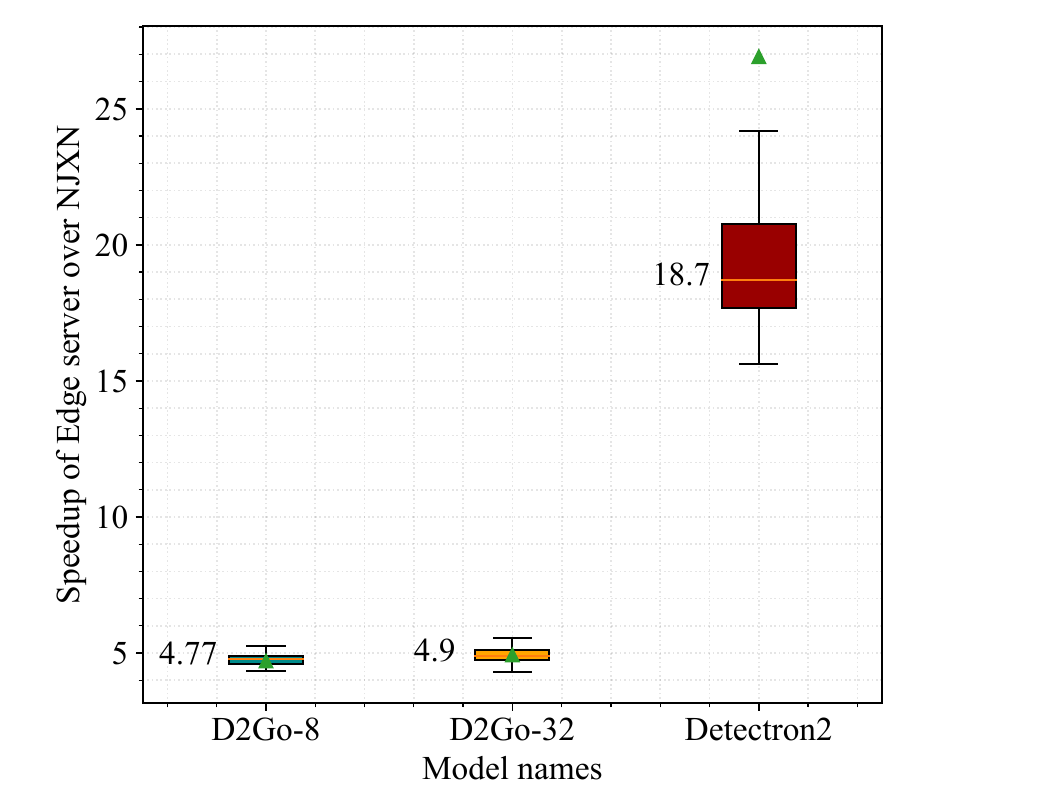}
\caption{\textbf{Class segmentation speedup per frame}. The distribution of the speedup of the edge-server over NJXN per frame for the test dataset. The results show that D2Go-8 and D2Go-32 have similar speedups, while Detectron2 has a median speedup of 18.7. }
\label{fig_pred_fps_dist_a}
\end{figure}

We compare the computational throughput of different models on the edge-server and the NJXN board. Fig.~\ref{fig_pred_fps_dist_a} shows a computation speedup in predicting frames for D2Go-8, D2Go-32 and Detectron2 of the edge-server over the NJXN board.

\noindent Our results manifest that the edge-server enables high-quality segmentation outputs, namely by Detectron2, with a median speedup of 18.7 over NJXN. 
We observe both D2Go models have a median speedup of just above four on the edge-server over NJXN. 
Robot controllers usually run above 1 kHz. This is called hard real-time. Soft real-time to implement impedance controller for interacting with hard objects is 100 Hz or above~\cite{augustyn2015real}. 
Biological latency observed in humans is 70 ms~\cite{cole1988grip}. Detectron2 and D2Go latency on the edge-server is almost 80 and 30 ms, respectively. That indicates we can obtain near-human performances in picking strawberries with future high-frequency dexterous manipulation controllers. 

Subject to available GPU resources on the edge-server, it can simultaneously provide segmentation service to many robots. Our system can handle three robots concurrently. We conclude that Detectron2 is suitable for semantic segmentation due to its enhanced quality, with the trade-off of slower processing on the edge-server.

\begin{figure}[tb!]\vspace{-3mm}
\centering
\includegraphics[scale=0.6]{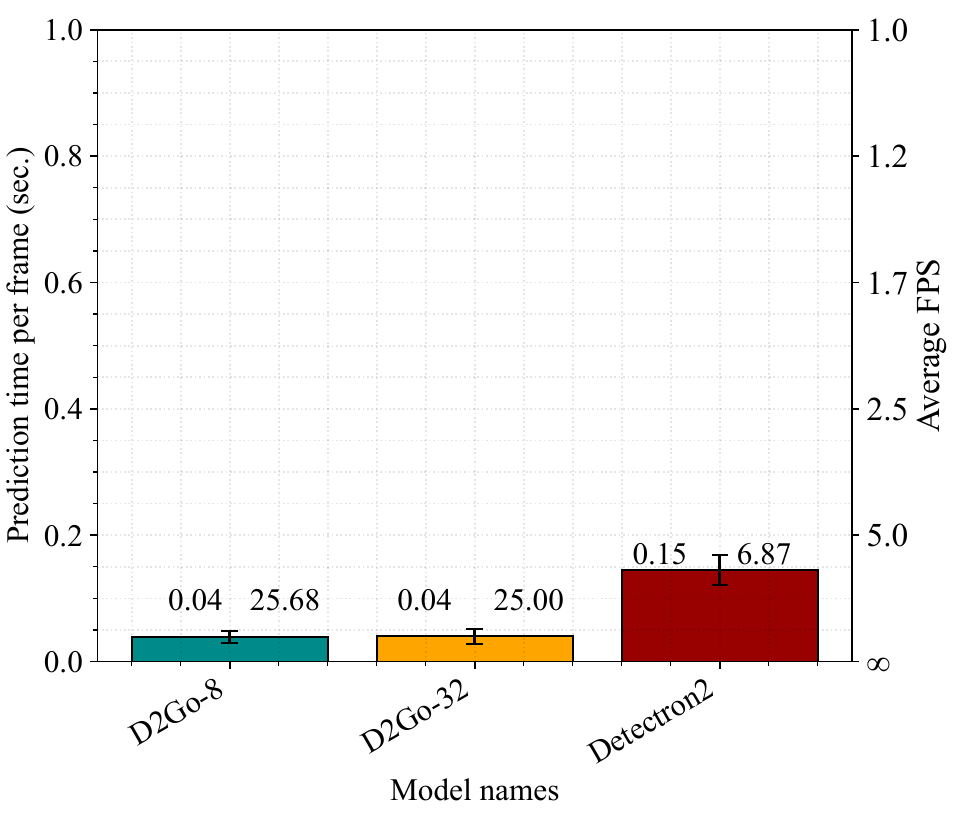}
\caption{ 
The distribution of the prediction time frame per second for the test dataset. The results show D2Go-8 and D2Go again are faster than Detectron2.}
\label{fig_pred_fps_dist_b}
\end{figure}
\subsection{Network Performance} 
\label{sec:results:network}
Several E5SH experiments were conducted at the Riseholme campus of the University of Lincoln on different days, times of day, climate conditions, etc. To analyse our system's communication efficiency and stability, we performed a comparative study on the performance of communication over 5G and WiFi networks. Below, we present our results in two categories, system latency and throughput speed. 

\begin{figure}[htpb!]\vspace{-3mm}
\centering
\includegraphics[scale=0.6]{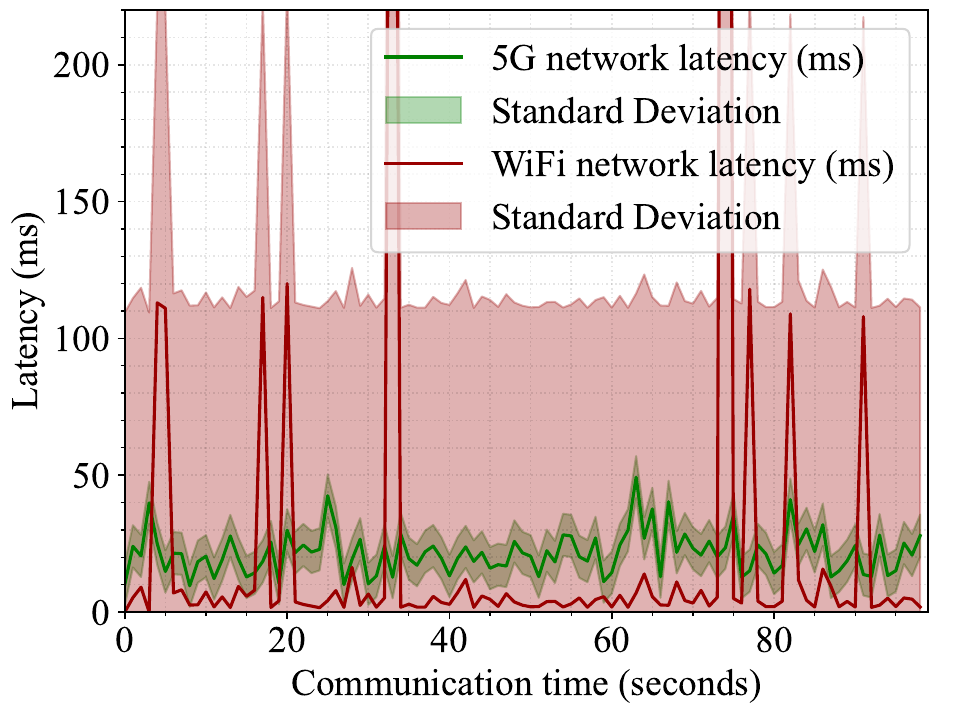}
\caption{\textbf{Network latency and communication time:} 5G  and WiFi networks.}
\label{fig_network_latency}
\end{figure}

Latency is measured by how fast a request message is sent, processed, and responded to on a network involving robots and an edge-server, denoted as Round-Trip Time (RTT). 
Reliability and consistency of message passing performance are as crucial as the Edge-server performance.
If wireless network throughput is not saturated, the throughput is not as critical to communication performance in terms of latency. In the E5SH system, image data transmission is bi-directional (i.e. robot-to-server and server-to-robot). Our results demonstrate that 5G yields lower but consistent latency across all experiment types. 
The latency mentioned above is good enough for video streaming at 60 FPS. 
The WiFi and 5G latency results are presented in Fig.~\ref{fig_network_latency}.

We experienced slightly poorer WiFi performance in the field, as indicated by significantly larger values of standard deviations compared to 5G. These results show that the WiFi signal is unsuitable for outdoor environments as it is primarily designed indoors. Multiple factors may contribute to the reduced performance of WiFi compared to indoor environments, such as obstacles, scaffolding and metal structures. Fig.~\ref{fig_network_latency} shows the edge-server's latency and indicates that 5G has a much larger capacity to stream higher FPS than WiFi. 

\subsection{Communication Factors}
\label{sec:results:comm-fa}
\begin{figure}[htpb]
\centering
\begin{subfigure}{0.4\textwidth}
  \centering
    \includegraphics[width=\textwidth]{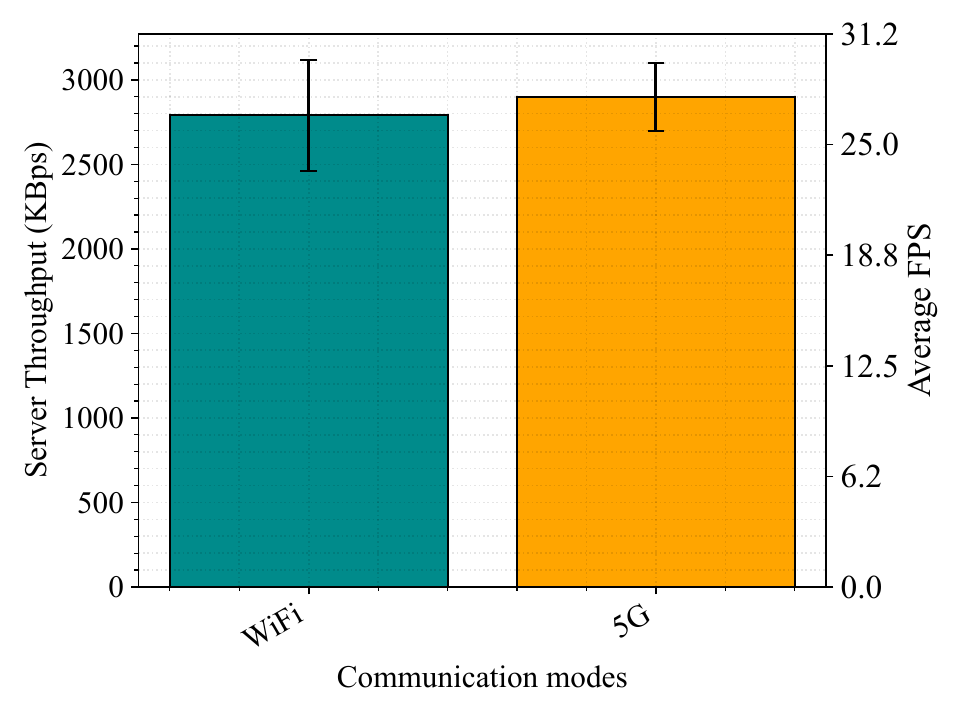}
  \label{fig:server_throughput_comm_modes}
  \caption{}
\end{subfigure}\
\begin{subfigure}{0.4\textwidth}
  \centering
    \includegraphics[width =\textwidth]{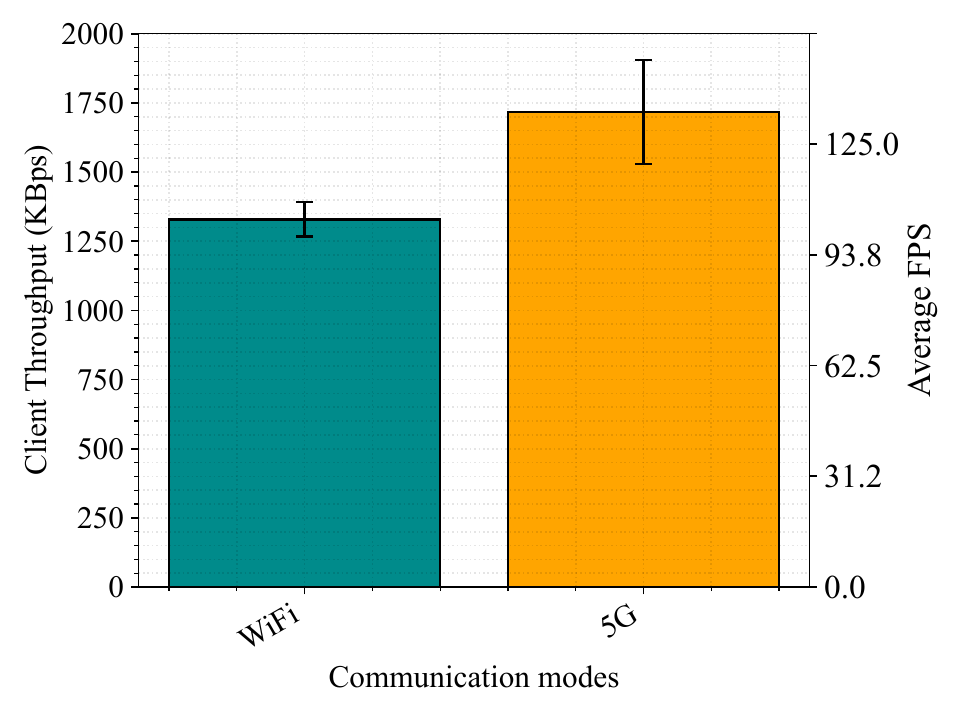}
  \label{fig:client_throughput_comm_modes}
  \caption{}
\end{subfigure}
\begin{subfigure}{0.4\textwidth}
  \centering

    \includegraphics[width =\textwidth]{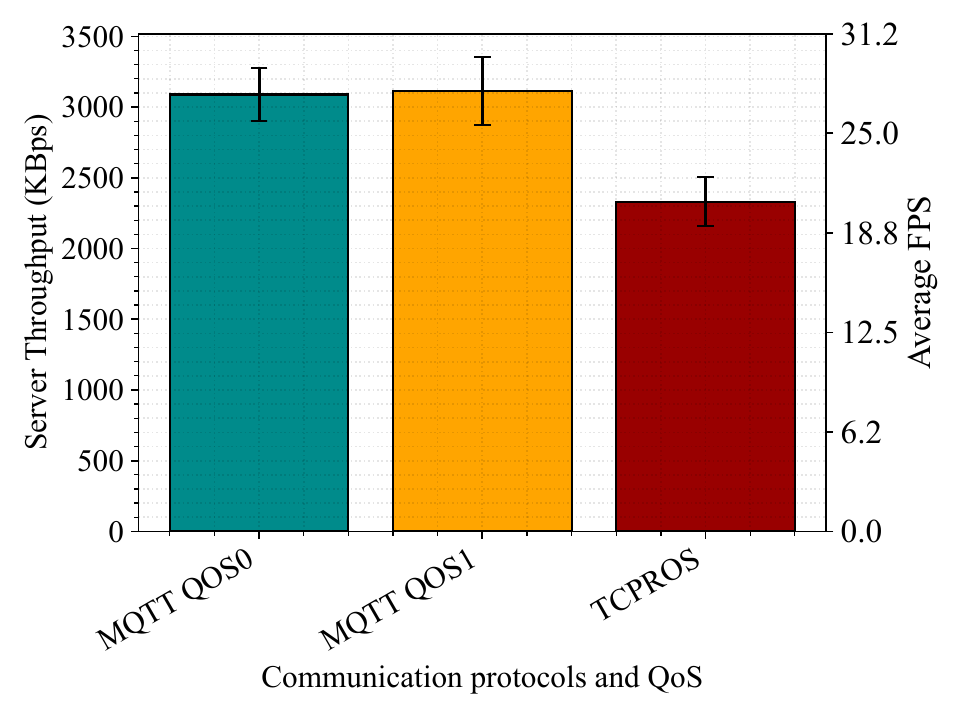}
  \label{fig:server_throughput_comm_protocols}
  \caption{}
\end{subfigure}
\begin{subfigure}{0.4\textwidth}
  \centering
    \includegraphics[width =\textwidth]{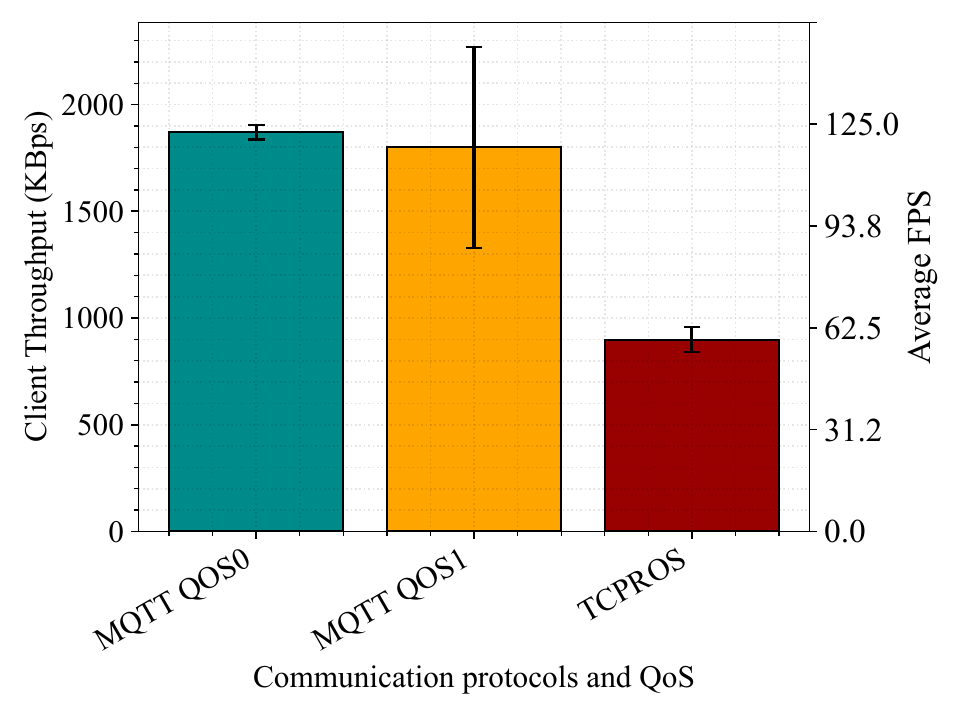}
  \label{fig:client_throughput_comm_protocols}
  \caption{}
\end{subfigure}
\begin{subfigure}{0.4\textwidth}
  \centering
    \includegraphics[width =\textwidth]{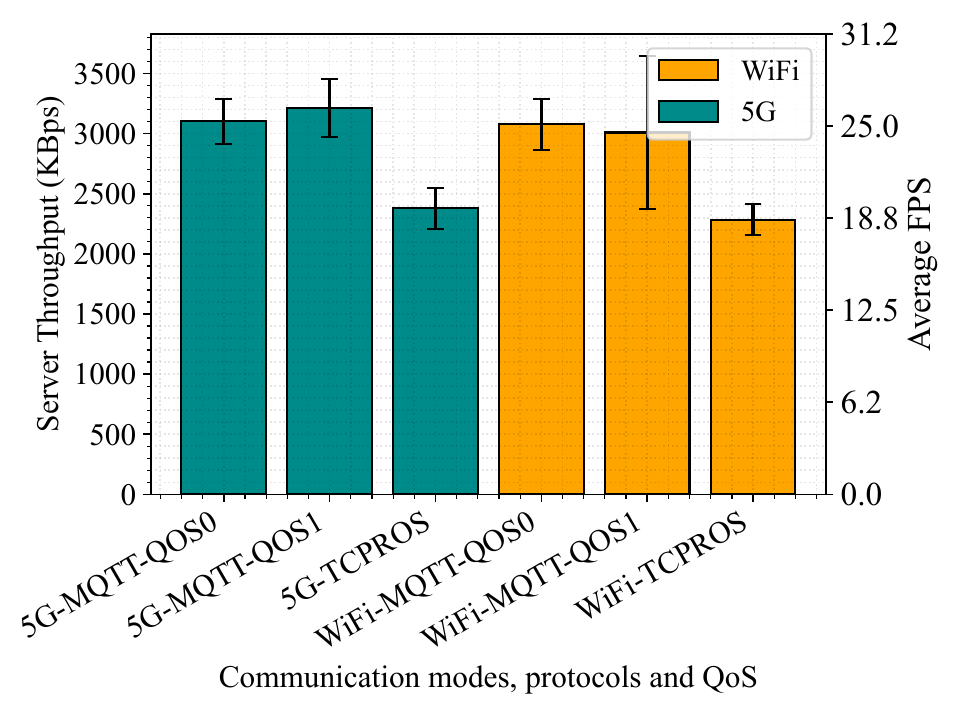}
  \label{fig:server_throughput}
  \caption{}
\end{subfigure}
\begin{subfigure}{0.4\textwidth}
  \centering
    \includegraphics[width =\textwidth]{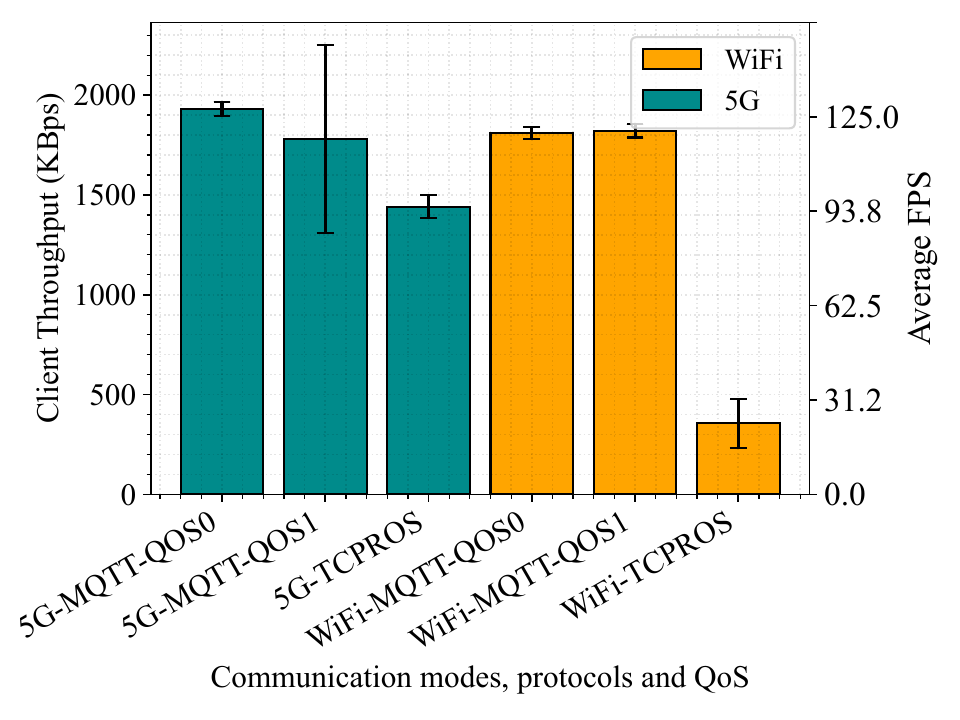}
  \label{fig:client_throughput}
  \caption{}
\end{subfigure}
    \caption{\textbf{Network throughput.} comparison of reception at the (a) server over WiFi and 5G networks. (b) client, over WiFi and 5G networks. (c) server, employing two protocols with QoS configurations. (d) client, employing two protocols with QoS configurations. (e) Robot-to-server communication. RGB-D transmitted having an average frame size of $80 KB$. (f) Server-to-robot communication. Segmented labels are transmitted having an average frame size of $16 KB$.}
    \label{fig_network_througput}
\end{figure}

We studied the throughput of different configurations through a stream of images with an average size of $80.0$ kB per image sent from the robot to the server over 5G and WiFi networks. 
After the edge-server segments images, it returns a stream of segmented images to the robot with an average size of $16$ kB.
We use small image sizes to remain below the network throughput saturation threshold for 5G and WiFi.
From the robot to the server, images are transmitted at 30 FPS. For instance, the throughput will be approximately 2400 KBps if each image is 80 kB. From the server to the robot, images are transmitted at 50 FPS. This results in a throughput of approximately 1250 KBps if each image is 25 kB. In both cases, we do not account for the overhead introduced by the transmission protocol. Since we send many packets of data simultaneously, we can analyse the performance of 5G and WiFi networks with different network protocols and throughput.
Moreover, we studied how they manage the traffic; the only change across different experiments was the type of wireless network. Hence, we expect similar results for 5G and WiFi, where 5G shows a higher transmission speed. The WiFi network shows lower upload and download throughput results and considerably lower performance when using TCPROS (Fig.~\ref{fig_network_througput}).

Fig.~\ref{fig_network_througput} shows network throughput data recorded at the edge-server and robots. It indicates that WiFi has a lower mean throughput, specifically in 
 WiFi TCPROS mode. The average frame transmission rate from the robot to the server (received throughput) is approximately 25 and 24.5 FPS with the 5G and WiFi networks, respectively. The average frame rate from the server to the robot (transmitted throughput) is 125 FPS over a 5G network and 120 FPS over WiFi. In this case, the higher FPS is because of the smaller size of segmented images· MQTT-based communication over 5G and WiFi outperforms other configurations  (Fig.~\ref{fig_network_througput}). Although 5G only marginally outperforms WiFi in terms of throughput (Fig.~\ref{fig_network_througput}), the latency results in Fig.~\ref{fig_network_latency} show 5G is significantly more stable than WiFi in outdoor environments.  
MQTT-QOS0 performs marginally better than MQTT-QOS1 in receiving data from the client, whereas the other differences between QOS0 and QOS1 are not statistically significant, as found when performing factor analysis (see below).
Hence, we use MQTT-QOS0 in subsequent sections and refer to it as 5G-MQTT or WiFi-MQTT.

We performed $2\times3$ factor analysis on the communication throughput data to assess the network infrastructure's impact (5G vs WiFi) in conjunction with the messaging protocol (MQTT-QOS0 vs MQTT-QOS1 vs TCPROS).
First, we test for statistical significance within each factor, as shown in the first four subfigures plotted in Figure~\ref{fig_network_througput}.
The results show statistically significant differences in performance for 5G vs WiFi at both the server and client, where 5G is the faster infrastructure, and MQTT is the faster protocol.
Note that the differences between MQTT-QOS0 and MQTT-QOS1 are not statistically significant.
The factor analysis shows that combining the two factors does not change the performance results for each factor taken individually.

\subsection{Cumulative System Speed-up} 
\begin{figure}[htb!]
\centering
\begin{subfigure}{0.49\textwidth}
  \centering
    \includegraphics[width=\textwidth]{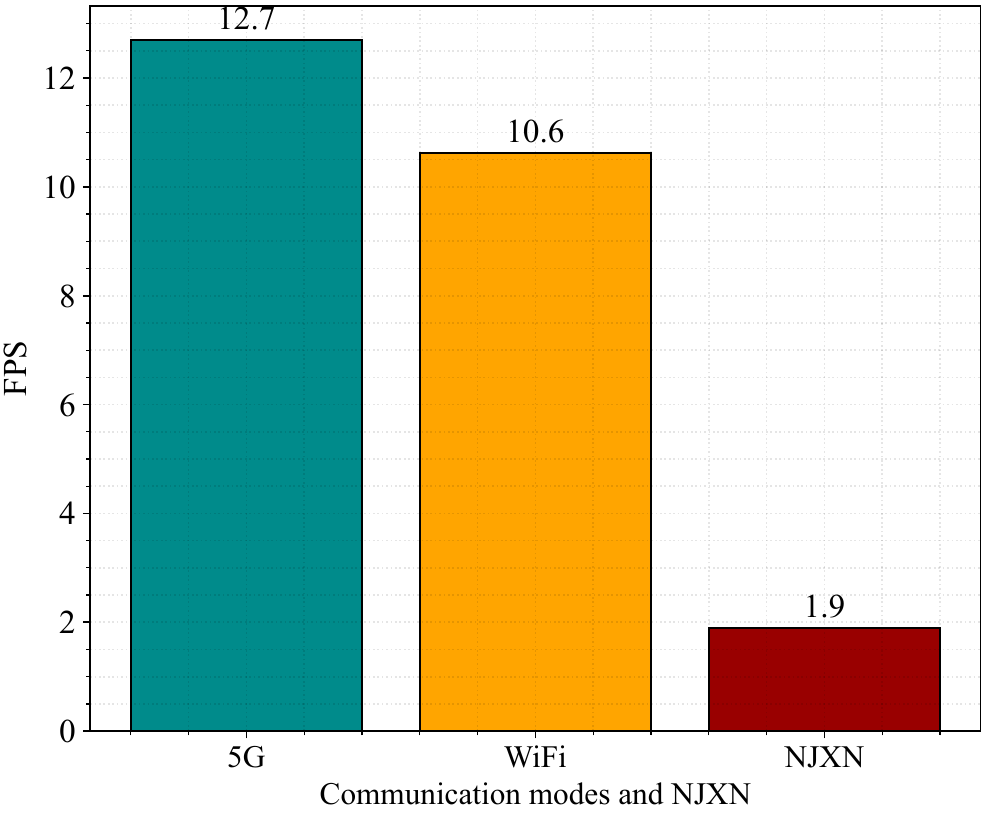}
  \label{fig:total_fps_comm_modes}
  \caption{}
\end{subfigure}\hfill
\begin{subfigure}{0.49\textwidth}
  \centering
    \includegraphics[width =\textwidth]{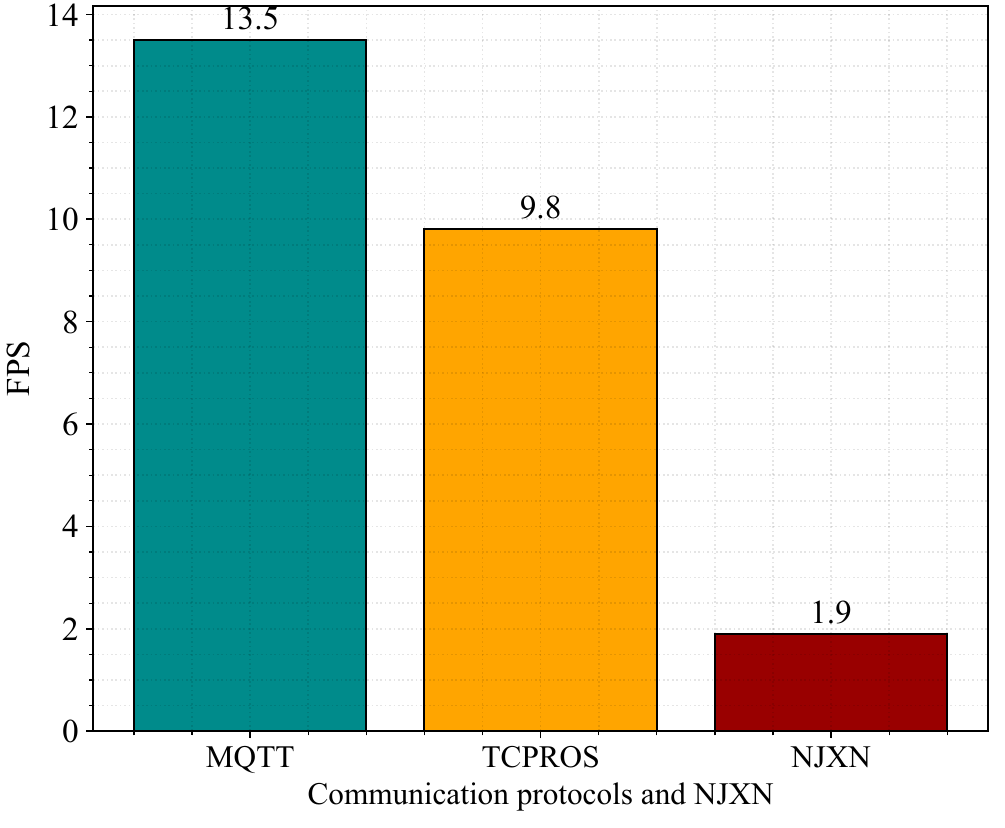}
  \label{fig:total_fps_comm_protocols}
  \caption{}
\end{subfigure}
\begin{subfigure}{0.47\textwidth}
  \centering
    \includegraphics[width =\textwidth]{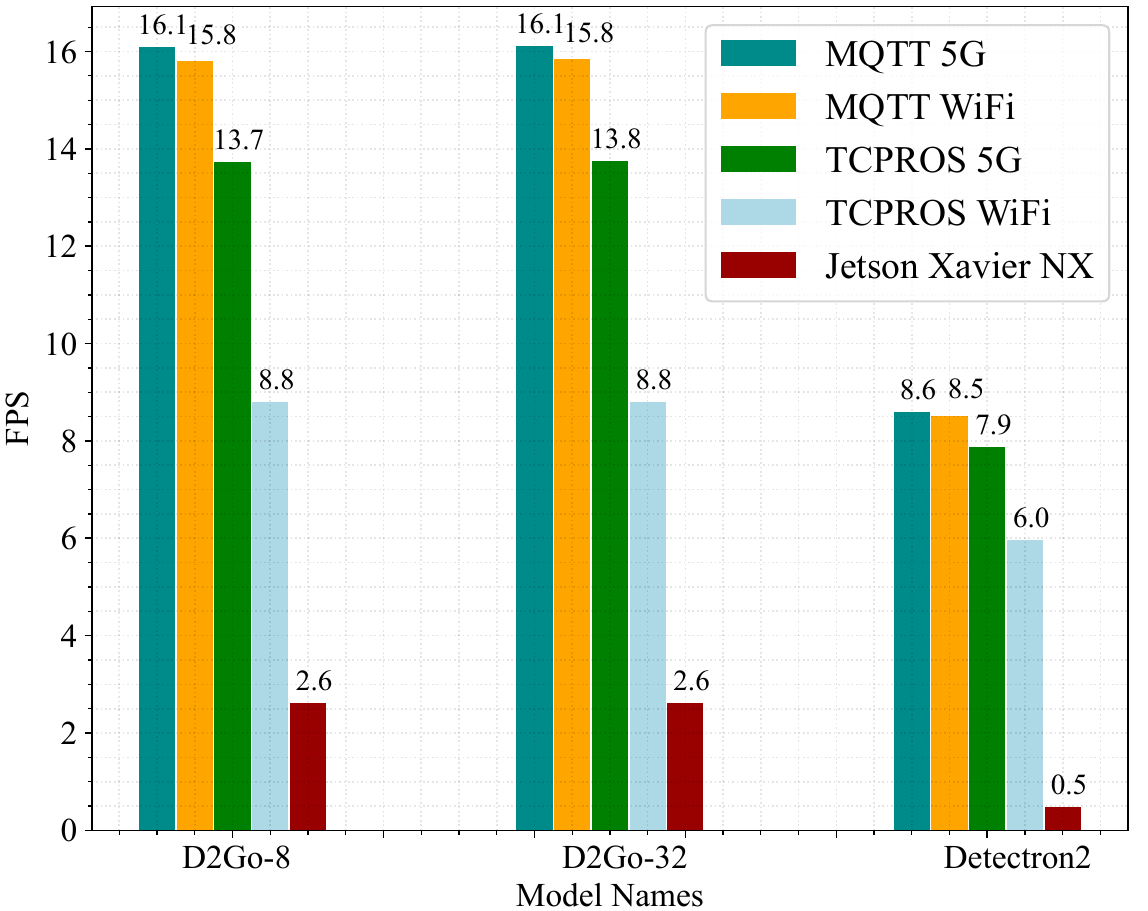}
  \label{fig:speedup1}
  \caption{}
\end{subfigure}
\begin{subfigure}{0.49\textwidth}
  \centering
    \includegraphics[width =\textwidth]{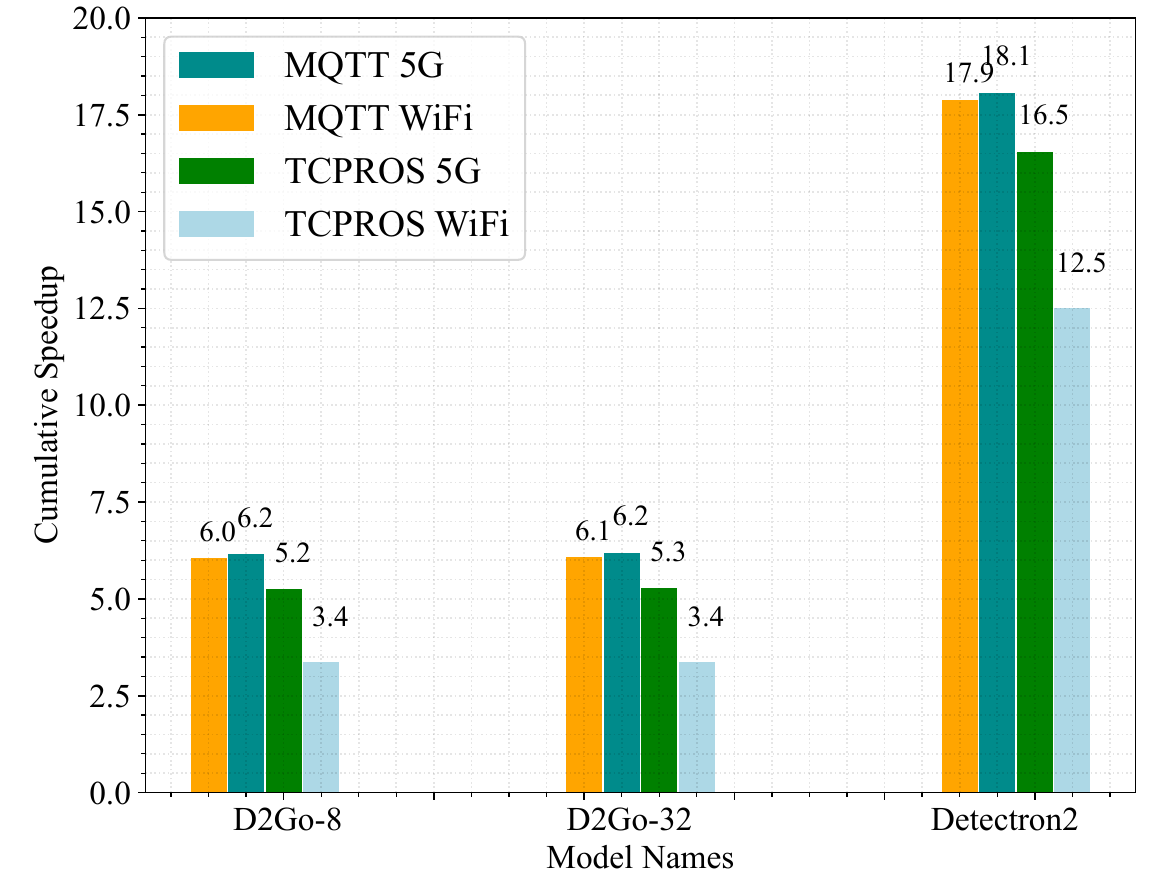}
  \caption{}
  \label{fig_speedup2}
\end{subfigure}
    \caption{(a) \textbf{Speedup and FPS} \textbf{for single robot} (a) Overall cumulative FPS for 5G, WiFi networks and NJXN (b) Overall cumulative FPS for MQTT, TCPROS and NJXN. (c) Overall cumulative FPS for 5G, WiFi networks, MQTT, TCPROS and NJXN for D2Go-8, D2Go-32 and Detectron2 models. (d) Overall cumulative Speedup for 5G, WiFi networks, MQTT, TCPROS and NJXN for D2Go-8, D2Go-32 and Detectron2 models. }
    \label{fig:cumulative-speedup}
\end{figure}

The segmentation, transmission, and reception delays during our tests helped us calculate the cumulative performance gain of our E5SH system over the embedded device. 
Figure~\ref{fig:cumulative-speedup}
shows the FPS performance of various configurations (e.g. (1) segmentation models on (2) NJXN board or using (3) MQTT and TCPROS communication protocols over (4) 5G or WiFi networks). We observe the maximum speed of our E5SH system with a setting including MQTT communication over the 5G network running the Detectron2 model, which is approximately 18.7 times faster than the embedded counterpart (Fig.~\ref{fig:cumulative-speedup}). Instead, D2Go-8 and D2Go-32 models are 4.8 and 4.9 times faster. In this case, Detectron2,  D2Go-8  and D2Go-32 on 5G-MQTT are 11.2, 5.3 and 5.2 times faster than the NJXN board, respectively.

\newpage
\section{Preliminary Analysis of Potential Future Work}

The centralised server-based segmentation opens several interesting aspects for comparison around the sustainability of our solution. In this section, we highlight comparative energy consumption, emission, scalability of rendering services to multiple robots and cost-benefit analysis of E5SH and NJXN systems. A supplementary comparative sustainability study was performed along with the core experiments of this work, which are presented in this section.  

\subsection{Energy consumption and emission}
 As energy consumption and $CO_2$ emissions are increasingly important, we also studied the sustainability aspect of our E5SH system. We computed the energy consumption of twelve robots based on the computations for one and three robots.
Here, we present power consumption and $CO_2$ emission per process by the package (called~\textit{energyusage})~\cite{Energyusage2022}. We considered the total emission of the United Kingdom \cite{Energyusage2022} as a reference. 

\begin{figure}[htb!]
\centering
\begin{subfigure}{0.49\textwidth}
  \centering
  \includegraphics[width=\textwidth]{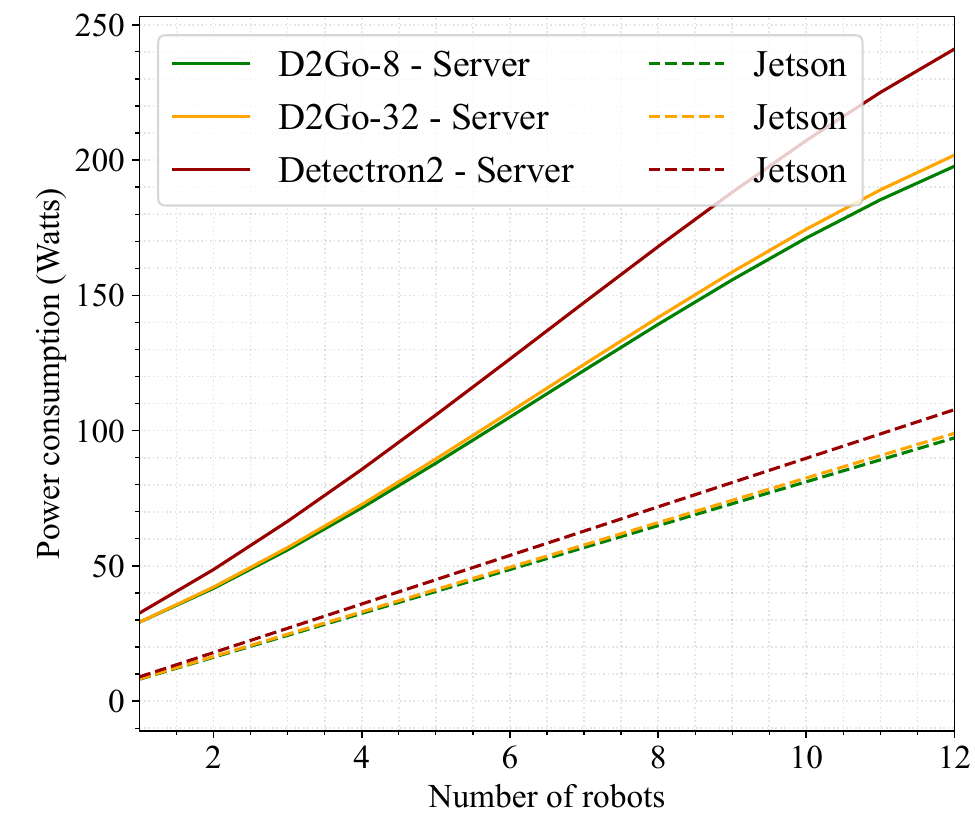}
  \caption{}
  \label{fig:energy_consumption_server}
\end{subfigure}
\hfill
\begin{subfigure}{0.49\textwidth}
  \centering
  \includegraphics[width=\textwidth]{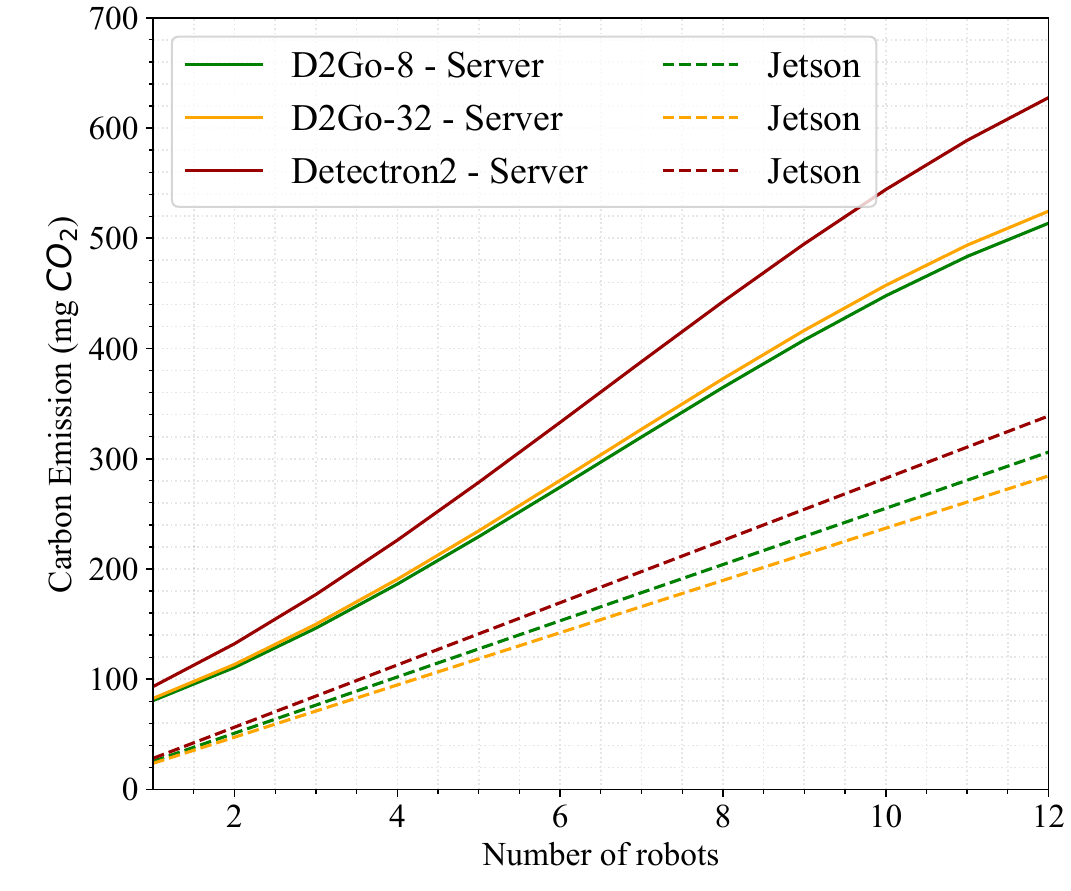}
  \caption{}
  \label{fig:emission_server}
\end{subfigure}
\caption{\textbf{\textbf{Power consumption analysis for multiple robots:}} (a) E5SH Power consumption (b) $CO_2$ emission; for using D2Go and Detectron2 on edge-server and NJXN board.}
    \label{fig_energy_and_co2_consumption}
\end{figure}

\begin{figure}[htpb!]
\centering
\begin{subfigure}{0.49\textwidth}
  \centering
  \includegraphics[width=\textwidth]{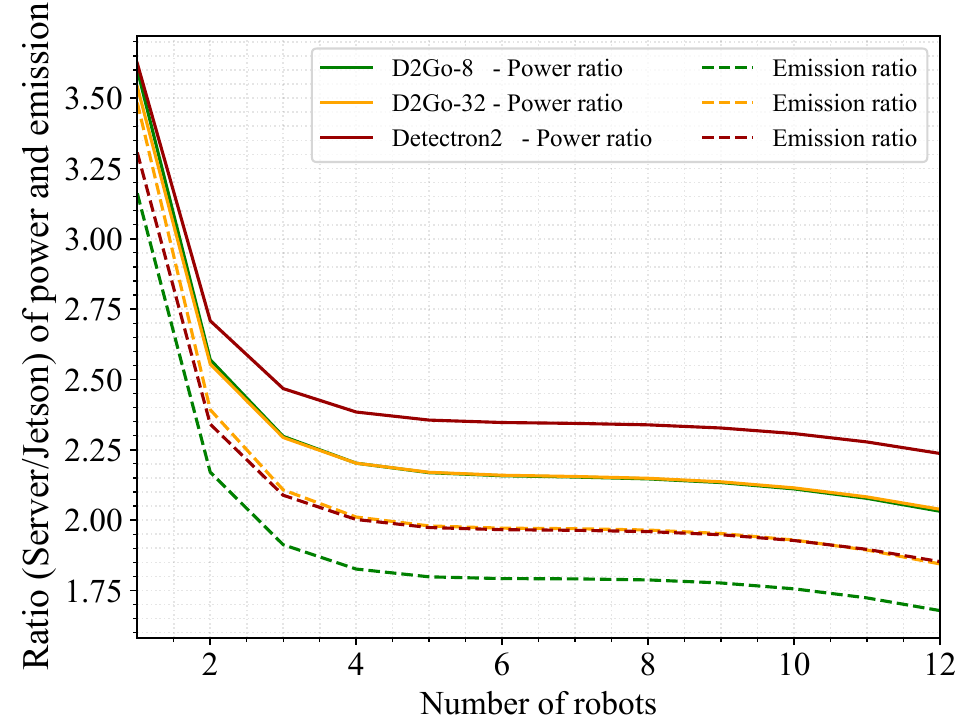}
  \caption{}

\end{subfigure}\hfill
\begin{subfigure}{0.49\textwidth}
  \centering

  \includegraphics[width=\textwidth]{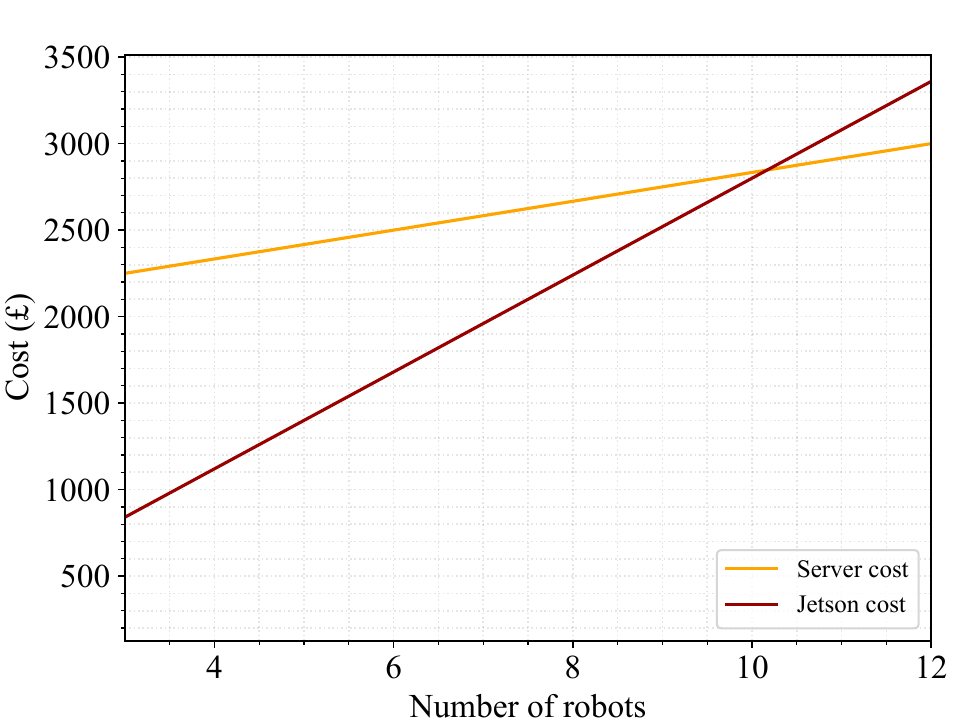}
  \caption{}

\end{subfigure}
\caption{Power consumption ratio and Cost-benefit analysis (On D2Go and Detectron2 on edge-server and NJXN board): (a) The ratio of power consumption to $CO_2$ emission for the edge-server and NJXN board; (b) cost-benefit for multiple robots using an edge-server and NJXN board.}
    \label{fig_cost_and_energy_ratio_consumption}
\end{figure}

Fig.~\ref{fig_energy_and_co2_consumption} shows the power consumption. On the edge-server, a single segmentation process running Detectron2 consumes 33.6 watts of power and emits 98 mg $CO_2$. Adding another process to serve the second robot results in 48.3 watts and 128 mg $CO_2$ energy use. 59.7 watts with 155 mg $CO_2$ are energy use and emission for the 3-robots case. For twelve robots, 240 watts and 200 watts are consumed for Detectron2 and D2Go, respectively, equivalent to 620 mg and 500 mg $CO_2$ emission. On the other hand, The power requirement reaches 110 watts in total when twelve robots have their standalone NJXN boards and are running Detectron2. This is approximately 95 watts for D2Go models. This is equivalent to 300 mg and 240 mg $CO_2$, respectively.
The trend shown in Fig.~\ref{fig_cost_and_energy_ratio_consumption} (a) confirms that the ratio reduces when the number of robots serviced by the edge-server increases. The ratio is above three for all models when one robot is served by one NJXN board and server, and it is under 2.4 for five robots.
\noindent Although NJXN boards are optimised for low energy use for onboard computing, increased robot energy use due to slower computing by NJXN board is another factor supporting the use of E5SH.  

We also conducted a cost-benefit analysis of computing devices. The edge-server's approximate recent cost is around £2000, and the GPU unit cost is approximately £250. The NJXN board costs around £300. Our multi-robot case study requires four GPUs but twelve NJXN boards to test twelve robots. Fig.~\ref{fig_cost_and_energy_ratio_consumption} (b) shows that the server setup is cost-efficient compared to standalone NJXN boards when ten or more robots are served. Additionally, maintaining and administering a fleet of robots with a standalone computing board is not scalable, and it is a commercially less viable solution. On the other hand, a failure of one NJXN board would affect only a single robot, but it will affect all twelve robots in case of server failure. 

In our proposed setup, we employed the edge-server only for semantic segmentation and performed the OctoMap generation on a robot-mounted laptop, which hosts the Moveit framework of ROS. The Moveit framework could also be deployed to the edge-server in the future to reduce the number of robot-mounted computing devices. Our edge-server configuration could be improved, and the GPU, such as Nvidia RTX-4090, could be installed to overcome the Detectron2 segmentation frame rate bottleneck. This could lead to greater than $20$ FPS performance. A test on a multi-server setup could also be conducted in the future to test the scaling potential in the real scenario.

\section{Discussion and conclusion}\vspace{-4mm} \label{sec_conclusion} 
Precision, speed, and computation cost are three factors relevant for a commercially viable selective harvesting robotic solution. 
Precise segmentation, 3-D localisation, and a complete scene map are computationally expensive. Precise segmentation models are usually more computationally demanding. As such, there is no scalable solution for near real-time manipulation actions. Cheaper and more effective energy-consuming computing boards are widely used to trade off the computation. 
We proposed an E5SH (edge-server over 5G selective harvesting) system enabling near real-time computations of segmentation, 3-D localisation, and motion planning for practical actions for strawberry picking. Our system transmits RGB-D images to the edge-server over a 5G network by employing MQTT and TCPROS communication protocols. After segmentation on the server, labels are returned to the robot for further action planning. Compared to our prior work~\cite{parsa2023autonomous}, which picks strawberry and puts it gently into a punnet in 25 seconds on average, E5SH can make the picking process almost 9 seconds faster, which is only 5 seconds away from 11 seconds, which is our human picking speed target. 

We conducted field tests with three different segmentation models and concluded that Detectron2 outperforms D2Go-8 and D2GO-32 regarding quality. D2Go yields inferior performance when segmenting rigid obstacles. This is a critical element of generating a collision map, as an imprecise collision estimation will damage the robot end-effector. Detectron2, in contrast, performs well in segmenting challenging obstacles. The quality of strawberry segmentation of Detectron2 is also better than that of D2Go; therefore, it is well suited for 3D localisation and collision map generation.
Nonetheless, Detectron2 needs heavy computation; hence, it is much slower on NJXN boards and unsuitable. Interestingly, Detectron2 on the edge-server segments 12.2 FPS. The network communication adds an overhead delay, which causes the drop in the average frame rate to 8.6, which is still an 18-fold gain compared to the NJXN board. 

We also conduct a multi-robot setup to quantify the carbon footprint and analyse the cost-benefit of deploying a server. The ratio of power consumption and respective carbon emission for the edge-server serving up to 12 robots to the case of robots with NJXN is decreasing with an increase in the number of robots. We see a sharp reduction in this ratio from one to four robots. Even though this ratio is about 2 for 12 robots, the hidden energy cost of having the robot still waiting for the computation to produce the action can be considered in future works. Furthermore, embedded devices are prone to shorter life since they benefit from less maintenance quality (e.g. exposure to weather change, moisture and shocks due to motion). We eventually show the cost-benefit of deploying a server solution that may serve up to twelve robots with a PC having four GPUs. If our server serves ten or more robots, the cost of having individual NJXN boards would surpass the server's cost. When three robots are served through a single GPU, the overall frame rate of the system drops to 5.7 from 8.6. It is worth noting that we can increase this frame rate by allocating more GPUs to the Edge computing machine. 

The network communication was performed over 5G and WiFi networks. Our network latency tests confirm that the instability in the 5G network is less than the WiFi counterpart, manifesting that 5G is a stable and well-suited mode of communication for harvesting operations. Two different protocols, MQTT and TCPROS, were tested using different QOS configurations for MQTT. The throughput results confirm that MQTT outperforms TCPROS in the network throughput both from the server and the client side. The QOS0 was the optimal configuration for MQTT communication. 
Edge computing and cloud computing are emerging technologies with future improvements in cost and speed. Our study provides a baseline to highlight the advantages of edge computing over 5G in agri-robotics. 

To deliver robots to human cost parity, the ultimate goal of crop harvesting robotic systems, we suggest that future robotic crop harvesting systems deployed at scale may require substantial network architecture, investment, and infrastructure.

The \emph{source code} developed during this project could be found at \cite{Fastpick2022a}. Machine learning models of Detectron2, Mask-RCNN and D2Go-32, D2Go-8 versions, field-tests videos and test datasets are shared at \cite{Fastpick2022b}.

\section{Acknowledgements}

The authors are grateful to the United Kingdom Research and Innovation (UKRI) for supporting this research through several grants. These grants are provided under UKRI Research England Lincoln Agri Robotics program, UKRI Research England CERES Agri-tech AI Unleashed program, and Robofruit, URKI Innovate UK Fastpick grant no. 99863 and UKRI EPSRC AgriForwards CDT. We are also thankful to the staff at the University of Lincoln, particularly Joni Appleton for the project management and Swati Megha, Luke Mahoney and Jonathan Trotter for facilitation during field tests.

\bibliographystyle{IEEEtran}
\bibliography{references}

\begin{thebibliography}{10}
\providecommand{\url}[1]{#1}
\csname url@samestyle\endcsname
\providecommand{\newblock}{\relax}
\providecommand{\bibinfo}[2]{#2}
\providecommand{\BIBentrySTDinterwordspacing}{\spaceskip=0pt\relax}
\providecommand{\BIBentryALTinterwordstretchfactor}{4}
\providecommand{\BIBentryALTinterwordspacing}{\spaceskip=\fontdimen2\font plus
\BIBentryALTinterwordstretchfactor\fontdimen3\font minus
  \fontdimen4\font\relax}
\providecommand{\BIBforeignlanguage}[2]{{%
\expandafter\ifx\csname l@#1\endcsname\relax
\typeout{** WARNING: IEEEtran.bst: No hyphenation pattern has been}%
\typeout{** loaded for the language `#1'. Using the pattern for}%
\typeout{** the default language instead.}%
\else
\language=\csname l@#1\endcsname
\fi
#2}}
\providecommand{\BIBdecl}{\relax}
\BIBdecl

\bibitem{pearson2022robotics}
S.~Pearson, C.~Camacho-Villa, R.~Valluru, G.~Oorbessy, M.~Rai, I.~Gould,
  S.~Brewer, E.~Sklar \emph{et~al.}, ``Robotics and autonomous systems for net
  zero agriculture,'' \emph{AGRICULTURE ROBOTICS Current Robotics Reports},
  vol.~3, pp. 57--64, 2022.

\bibitem{duckett:agri-robots}
\BIBentryALTinterwordspacing
T.~Duckett, S.~Pearson, S.~Blackmore, B.~Grieve, W.-H. Chen, G.~Cielniak,
  J.~Cleaversmith, J.~Dai, S.~Davis, C.~Fox, P.~From, I.~Georgilas, R.~Gill,
  I.~Gould, M.~Hanheide, A.~Hunter, F.~Iida, L.~Mihalyova, S.~Nefti-Meziani,
  G.~Neumann, P.~Paoletti, T.~Pridmore, D.~Ross, M.~Smith, M.~Stoelen,
  M.~Swainson, S.~Wane, P.~Wilson, I.~Wright, and G.-Z. Yang, ``Agricultural
  robotics: The future of robotic agriculture,'' 2018. [Online]. Available:
  \url{https://arxiv.org/abs/1806.06762}
\BIBentrySTDinterwordspacing

\bibitem{marinoudi2019robotics}
V.~Marinoudi, C.~G. S{\o}rensen, S.~Pearson, and D.~Bochtis, ``Robotics and
  labour in agriculture. a context consideration,'' \emph{Biosystems
  Engineering}, vol. 184, pp. 111--121, 2019.

\bibitem{marinoudi2021future}
V.~Marinoudi, M.~Lampridi, D.~Kateris, S.~Pearson, C.~G. S{\o}rensen, and
  D.~Bochtis, ``The future of agricultural jobs in view of robotization,''
  \emph{Sustainability}, vol.~13, no.~21, p. 12109, 2021.

\bibitem{Bloomberg2022}
M.~Durisin, ``Uk worker shortage leaves £60 million of food to rot in
  fields,''
  \url{https://www.bloomberg.com/news/articles/2022-08-15/uk-worker-shortage-leaves-60-million-of-food-to-rot-in-fields},
  2022.

\bibitem{Eatingwell2022}
B.~Estabrook, ``Farmers can't find enough workers to harvest crops,''
  \url{https://lncn.ac/shortlabor}, 2022.

\bibitem{EastAsiaForum2022}
Y.~Yoshikawa, ``Combatting japan’s agricultural worker shortage,''
  \url{https://www.eastasiaforum.org/2022/03/03/combatting-japans-agricultural-worker-shortage/},
  2022.

\bibitem{HarvestCroo2022}
\BIBentryALTinterwordspacing
M.~Harvest CROO Robotics~Team, ``{Harvest CROO Robotics Homepage},'' 2022.
  [Online]. Available: \url{https://www.harvestcroorobotics.com/}
\BIBentrySTDinterwordspacing

\bibitem{Zhang2023}
\BIBentryALTinterwordspacing
K.~Zhang, K.~Lammers, P.~Chu, Z.~Li, and R.~Lu, ``An automated apple harvesting
  robot—from system design to field evaluation,'' \emph{Journal of Field
  Robotics}, vol. n/a, no. n/a, 2023. [Online]. Available:
  \url{https://onlinelibrary.wiley.com/doi/abs/10.1002/rob.22268}
\BIBentrySTDinterwordspacing

\bibitem{Xiong2022}
\BIBentryALTinterwordspacing
Z.~Xiong, Q.~Feng, T.~Li, F.~Xie, C.~Liu, L.~Liu, X.~Guo, and C.~Zhao,
  ``Dual-manipulator optimal design for apple robotic harvesting,''
  \emph{Agronomy}, vol.~12, no.~12, 2022. [Online]. Available:
  \url{https://www.mdpi.com/2073-4395/12/12/3128}
\BIBentrySTDinterwordspacing

\bibitem{Metomotion2022}
W.~Metomotion~homepage, ``Metomotion harvesting robots,''
  \url{https://metomotion.com/}, 2022.

\bibitem{Mohanan2021}
\BIBentryALTinterwordspacing
M.~G. Mohanan and A.~Salgaonkar, ``Robotic mushroom harvesting by employing
  probabilistic road map and inverse kinematics,'' \emph{Bohr International
  Journal of Furutre Robotics and Artificial Intelligence}, vol.~1, no.~1,
  2021. [Online]. Available:
  \url{https://www.bohrpub.com/journals/BIJFRAI/Vol1N1/BIJFRAI_20211101}
\BIBentrySTDinterwordspacing

\bibitem{Blok2016}
\BIBentryALTinterwordspacing
P.~M. Blok, R.~Barth, and W.~{van den Berg}, ``Machine vision for a selective
  broccoli harvesting robot,'' \emph{IFAC-PapersOnLine}, vol.~49, no.~16, pp.
  66--71, 2016, 5th IFAC Conference on Sensing, Control and Automation
  Technologies for Agriculture AGRICONTROL 2016. [Online]. Available:
  \url{https://www.sciencedirect.com/science/article/pii/S2405896316315749}
\BIBentrySTDinterwordspacing

\bibitem{Li2023}
T.~Li, F.~Xie, Q.~Qiu, and Q.~Feng, ``Multi-arm robot task planning for fruit
  harvesting using multi-agent reinforcement learning,'' in \emph{2023 IEEE/RSJ
  International Conference on Intelligent Robots and Systems (IROS)}, 2023, pp.
  4176--4183.

\bibitem{Mann2016}
\BIBentryALTinterwordspacing
M.~P. Mann, B.~Zion, I.~Shmulevich, D.~Rubinstein, and R.~Linker,
  ``Combinatorial optimization and performance analysis of a multi-arm
  cartesian robotic fruit harvester---extensions of graph coloring,''
  \emph{Journal of Intelligent {\&} Robotic Systems}, vol.~82, no.~3, pp.
  399--411, Jun 2016. [Online]. Available:
  \url{https://doi.org/10.1007/s10846-015-0211-5}
\BIBentrySTDinterwordspacing

\bibitem{LuisaChesshire2022}
\BIBentryALTinterwordspacing
L.~Chesshire, ``Dogtooth takes robots to downing street,'' \emph{Fresh Produce
  Journal}, 2022. [Online]. Available:
  \url{https://www.fruitnet.com/fresh-produce-journal/dogtooth-takes-robots-to-downing-street/246308.article}
\BIBentrySTDinterwordspacing

\bibitem{Ge2020}
\BIBentryALTinterwordspacing
Y.~Ge, Y.~Xiong, and P.~From, ``Symmetry-based 3d shape completion for fruit
  localisation for harvesting robots,'' \emph{Biosystems Engineering}, vol.
  197, pp. 188--202, 2020. [Online]. Available:
  \url{https://www.sciencedirect.com/science/article/pii/S1537511020301963}
\BIBentrySTDinterwordspacing

\bibitem{Agrobotics2022}
\BIBentryALTinterwordspacing
M.~Agro Robotics~Team, ``Agro robotics homepage,'' 2022. [Online]. Available:
  \url{https://www.agrobot.com/e-series}
\BIBentrySTDinterwordspacing

\bibitem{Octinion2018}
\BIBentryALTinterwordspacing
A.~{De Preter}, J.~Anthonis, and J.~{De Baerdemaeker}, ``Development of a robot
  for harvesting strawberries,'' \emph{IFAC-PapersOnLine}, vol.~51, no.~17, pp.
  14--19, 2018, 6th IFAC Conference on Bio-Robotics BIOROBOTICS 2018. [Online].
  Available:
  \url{https://www.sciencedirect.com/science/article/pii/S2405896318311704}
\BIBentrySTDinterwordspacing

\bibitem{Certhon2024}
W.~Certhon~homepage, ``Certhon harvesting robots,''
  \url{https://certhonharvestrobot.com/}, 2024.

\bibitem{FourGrowers}
W.~Four Growers~homepage, ``Four growers robotic harvesting and plant
  analytics,'' \url{https://fourgrowers.com/}, 2024.

\bibitem{Mehta2022}
\BIBentryALTinterwordspacing
S.~Mehta, W.~MacKunis, and T.~Burks, ``Robust visual servo control in the
  presence of fruit motion for robotic citrus harvesting,'' \emph{Computers and
  Electronics in Agriculture}, vol. 123, pp. 362--375, 2016. [Online].
  Available:
  \url{https://www.sciencedirect.com/science/article/pii/S0168169916300746}
\BIBentrySTDinterwordspacing

\bibitem{Mehta2014}
\BIBentryALTinterwordspacing
S.~Mehta and T.~Burks, ``Vision-based control of robotic manipulator for citrus
  harvesting,'' \emph{Computers and Electronics in Agriculture}, vol. 102, pp.
  146--158, 2014. [Online]. Available:
  \url{https://www.sciencedirect.com/science/article/pii/S0168169914000052}
\BIBentrySTDinterwordspacing

\bibitem{Luo2018}
\BIBentryALTinterwordspacing
L.~Luo, Y.~Tang, Q.~Lu, X.~Chen, P.~Zhang, and X.~Zou, ``A vision methodology
  for harvesting robot to detect cutting points on peduncles of double
  overlapping grape clusters in a vineyard,'' \emph{Computers in Industry},
  vol.~99, pp. 130--139, 2018. [Online]. Available:
  \url{https://www.sciencedirect.com/science/article/pii/S0166361517305298}
\BIBentrySTDinterwordspacing

\bibitem{Barnett2020}
\BIBentryALTinterwordspacing
J.~Barnett, M.~Duke, C.~K. Au, and S.~H. Lim, ``Work distribution of multiple
  cartesian robot arms for kiwifruit harvesting,'' \emph{Computers and
  Electronics in Agriculture}, vol. 169, p. 105202, 2020. [Online]. Available:
  \url{https://www.sciencedirect.com/science/article/pii/S016816991931258X}
\BIBentrySTDinterwordspacing

\bibitem{Suraj2017}
\BIBentryALTinterwordspacing
S.~Amatya, M.~Karkee, Q.~Zhang, and M.~D. Whiting, ``Automated detection of
  branch shaking locations for robotic cherry harvesting using machine
  vision,'' \emph{Robotics}, vol.~6, no.~4, 2017. [Online]. Available:
  \url{https://www.mdpi.com/2218-6581/6/4/31}
\BIBentrySTDinterwordspacing

\bibitem{parsa2023autonomous}
S.~Parsa, B.~Debnath, M.~A. Khan, and A.~Ghalamzan, ``Modular autonomous
  strawberry picking robotic system,'' \emph{Journal of Field Robotics}, 2023.

\bibitem{rajendran2023towards}
V.~Rajendran, B.~Debnath, S.~Mghames, W.~Mandil, S.~Parsa, S.~Parsons, and
  A.~Ghalamzan, ``Towards autonomous selective harvesting: A review of robot
  perception, robot design, motion planning and control,'' \emph{Journal of
  Field Robotics}, 2023.

\bibitem{Xiong2018}
\BIBentryALTinterwordspacing
Y.~Xiong, Y.~Ge, L.~Grimstad, and P.~J. From, ``An autonomous
  strawberry-harvesting robot: Design, development, integration, and field
  evaluation,'' \emph{Journal of Field Robotics}, vol.~37, no.~2, pp. 202--224,
  2020. [Online]. Available:
  \url{https://onlinelibrary.wiley.com/doi/abs/10.1002/rob.21889}
\BIBentrySTDinterwordspacing

\bibitem{agronomy2022}
\BIBentryALTinterwordspacing
J.~K. Basak, B.~Paudel, N.~E. Kim, N.~C. Deb, B.~G. Kaushalya~Madhavi, and
  H.~T. Kim, ``Non-destructive estimation of fruit weight of strawberry using
  machine learning models,'' \emph{Agronomy}, vol.~12, no.~10, 2022. [Online].
  Available: \url{https://www.mdpi.com/2073-4395/12/10/2487}
\BIBentrySTDinterwordspacing

\bibitem{Ge2019b}
Y.~Ge, Y.~Xiong, G.~L. Tenorio, and P.~J. From, ``Fruit localization and
  environment perception for strawberry harvesting robots,'' \emph{IEEE
  Access}, vol.~7, 2019.

\bibitem{cole1988grip}
K.~J. Cole and J.~H. Abbs, ``Grip force adjustments evoked by load force
  perturbations of a grasped object,'' \emph{Journal of neurophysiology},
  vol.~60, no.~4, pp. 1513--1522, 1988.

\bibitem{tafuro2022strawberry}
A.~Tafuro, A.~Adewumi, S.~Parsa, A.~Ghalamzan, and B.~Debnath, ``Strawberry
  picking point localization ripeness and weight estimation,'' in \emph{2022
  International Conference on Robotics and Automation (ICRA)}.\hskip 1em plus
  0.5em minus 0.4em\relax IEEE, 2022, pp. 2295--2302.

\bibitem{hornung2013octomap}
A.~Hornung, K.~M. Wurm, M.~Bennewitz, C.~Stachniss, and W.~Burgard, ``Octomap:
  An efficient probabilistic 3d mapping framework based on octrees,''
  \emph{Autonomous robots}, vol.~34, no.~3, pp. 189--206, 2013.

\bibitem{Motionplan2022}
\BIBentryALTinterwordspacing
T.~Sandakalum and M.~H. Ang, ``Motion planning for mobile manipulators;a
  systematic review,'' \emph{Machines}, vol.~10, no.~2, 2022. [Online].
  Available: \url{https://www.mdpi.com/2075-1702/10/2/97}
\BIBentrySTDinterwordspacing

\bibitem{tafuro2022dpmp}
A.~Tafuro, B.~Debnath, A.~M. Zanchettin, and E.~A. Ghalamzan, ``dpmp-deep
  probabilistic motion planning: A use case in strawberry picking robot,'' in
  \emph{2022 IEEE/RSJ international conference on intelligent robots and
  systems (IROS)}.\hskip 1em plus 0.5em minus 0.4em\relax IEEE, 2022, pp.
  8675--8681.

\bibitem{sanni2022deep}
O.~Sanni, G.~Bonvicini, M.~A. Khan, P.~C. L{\'o}pez-Custodio, K.~Nazari
  \emph{et~al.}, ``Deep movement primitives: toward breast cancer examination
  robot,'' in \emph{Proceedings of the AAAI Conference on Artificial
  Intelligence}, vol.~36, no.~11, 2022, pp. 12\,126--12\,134.

\bibitem{mandil2023tactile}
W.~Mandil, V.~Rajendran, and K.~Nazari, ``Tactile-sensing technologies: Trends,
  challenges and outlook in agri-food manipulation,'' \emph{Sensors}, vol.~23,
  no.~17, p. 7362, 2023.

\bibitem{parsons2023acoustic}
V.~Rajendran, S.~Parsons, and A.~Ghalamzan, ``Acoustic soft tactile skin:
  Towards continuous tactile sensing,'' in \emph{2023 21st International
  Conference on Advanced Robotics (ICAR)}.\hskip 1em plus 0.5em minus
  0.4em\relax IEEE, 2023, pp. 621--626.

\bibitem{rajendran2024acoustic}
V.~Rajendran, S.~Parsons \emph{et~al.}, ``Acoustic soft tactile skin (ast),''
  \emph{IEEE International conference on robotics and automation (ICRA)}, 2024.

\bibitem{rajendran2024ast}
V.~Rajendran, S.~Parsons, and A.~Ghalamzan, ``Ast-2: Single and bi-layered 2-d
  acoustic soft tactile skin,'' \emph{IEEE Robosoft conference
  arXiv:2401.14292}, 2024.

\bibitem{rajendran202enabling}
V.~Rajendran, K.~Nazari, S.~Parsons, and A.~Ghalamzan, ``Enabling tactile
  feedback for robotic strawberry handling using ast skin,'' \emph{Towards
  Autonomous Robotic Systems: 25nd Annual Conference (TAROS)}, 2024.

\bibitem{vishnu2023acoustic}
W.~Mandil, S.~Parsons \emph{et~al.}, ``Acoustic soft tactile skin (ast skin),''
  \emph{IEEE International Conference on Robotics and Automation (ICRA)}, 2024.

\bibitem{nazari2023deep}
K.~Nazari, G.~Gandolfi, Z.~Talebpour, V.~Rajendran, W.~Mandil, P.~Rocco, and
  A.~Ghalamzan-E, ``Deep functional predictive control (deep-fpc): Robot
  pushing 3-d cluster using tactile prediction,'' in \emph{2023 IEEE/RSJ
  International Conference on Intelligent Robots and Systems (IROS)}.\hskip 1em
  plus 0.5em minus 0.4em\relax IEEE, 2023, pp. 10\,771--10\,776.

\bibitem{Ge2019a}
\BIBentryALTinterwordspacing
Y.~Ge, Y.~Xiong, and P.~J. From, ``Instance segmentation and localization of
  strawberries in farm conditions for automatic fruit harvesting,''
  \emph{IFAC-PapersOnLine}, vol.~52, no.~30, pp. 294--299, 2019, 6th IFAC
  Conference on Sensing, Control and Automation Technologies for Agriculture
  AGRICONTROL 2019. [Online]. Available:
  \url{https://www.sciencedirect.com/science/article/pii/S2405896319324565}
\BIBentrySTDinterwordspacing

\bibitem{Onishi2019}
\BIBentryALTinterwordspacing
Y.~Onishi, T.~Yoshida, H.~Kurita, T.~Fukao, H.~Arihara, and A.~Iwai, ``An
  automated fruit harvesting robot by using deep learning,'' \emph{ROBOMECH
  Journal}, vol.~6, no.~1, p.~13, Nov 2019. [Online]. Available:
  \url{https://doi.org/10.1186/s40648-019-0141-2}
\BIBentrySTDinterwordspacing

\bibitem{Williams2020}
\BIBentryALTinterwordspacing
H.~Williams, C.~Ting, M.~Nejati, M.~H. Jones, N.~Penhall, J.~Lim, M.~Seabright,
  J.~Bell, H.~S. Ahn, A.~Scarfe, M.~Duke, and B.~MacDonald, ``Improvements to
  and large-scale evaluation of a robotic kiwifruit harvester,'' \emph{Journal
  of Field Robotics}, vol.~37, no.~2, pp. 187--201, 2020. [Online]. Available:
  \url{https://onlinelibrary.wiley.com/doi/abs/10.1002/rob.21890}
\BIBentrySTDinterwordspacing

\bibitem{Ge2022}
\BIBentryALTinterwordspacing
Y.~Ge, Y.~Xiong, and P.~J. From, ``Three-dimensional location methods for the
  vision system of strawberry-harvesting robots: development and comparison,''
  \emph{Precision Agriculture}, Nov 2022. [Online]. Available:
  \url{https://doi.org/10.1007/s11119-022-09974-4}
\BIBentrySTDinterwordspacing

\bibitem{mask_rcnn2017}
K.~He, G.~Gkioxari, P.~Dollár, and R.~Girshick, ``Mask r-cnn,'' in \emph{2017
  IEEE International Conference on Computer Vision (ICCV)}, 2017, pp.
  2980--2988.

\bibitem{Xiong2020}
\BIBentryALTinterwordspacing
Y.~Xiong, Y.~Ge, and P.~J. From, ``An obstacle separation method for robotic
  picking of fruits in clusters,'' \emph{Computers and Electronics in
  Agriculture}, vol. 175, p. 105397, 2020. [Online]. Available:
  \url{https://www.sciencedirect.com/science/article/pii/S0168169919324214}
\BIBentrySTDinterwordspacing

\bibitem{Uijlings2013}
\BIBentryALTinterwordspacing
J.~R.~R. Uijlings, K.~E.~A. van~de Sande, T.~Gevers, and A.~W.~M. Smeulders,
  ``Selective search for object recognition,'' \emph{International Journal of
  Computer Vision}, vol. 104, no.~2, pp. 154--171, Sep 2013. [Online].
  Available: \url{https://doi.org/10.1007/s11263-013-0620-5}
\BIBentrySTDinterwordspacing

\bibitem{FBNet32021}
e.~a. X.~Dai, ``Fbnetv3: Joint architecture-recipe search using predictor
  pretraining,'' in \emph{Conference on Computer Vision and Pattern
  Recognition}.\hskip 1em plus 0.5em minus 0.4em\relax IEEE, 2021.

\bibitem{HastyAI2022}
HastAI, ``Fbnetv3: Model architecture wiki, hastyai,''
  \url{https://hasty.ai/docs/mp-wiki/model-architectures/fbnetv3}, 2022.

\bibitem{quigley2009ros}
M.~Quigley, K.~Conley, B.~Gerkey, J.~Faust, T.~Foote, J.~Leibs, R.~Wheeler,
  A.~Y. Ng \emph{et~al.}, ``Ros: an open-source robot operating system,'' in
  \emph{ICRA workshop on open source software}, vol.~3.\hskip 1em plus 0.5em
  minus 0.4em\relax Kobe, Japan, 2009, p.~5.

\bibitem{hornung13auro}
\BIBentryALTinterwordspacing
A.~Hornung, K.~M. Wurm, M.~Bennewitz, C.~Stachniss, and W.~Burgard,
  ``{OctoMap}: An efficient probabilistic {3D} mapping framework based on
  octrees,'' \emph{Autonomous Robots}, 2013, software available at
  \url{https://octomap.github.io}. [Online]. Available:
  \url{https://octomap.github.io}
\BIBentrySTDinterwordspacing

\bibitem{van20225g}
M.~van Hilten and S.~Wolfert, ``5g in agri-food-a review on current status,
  opportunities and challenges,'' \emph{Computers and Electronics in
  Agriculture}, vol. 201, p. 107291, 2022.

\bibitem{zhivkov:5g:ukras}
T.~Zhivkov, A.~Gomez, J.~Gao, E.~Sklar, and S.~Parsons, ``{The need for speed:
  How 5G communication can support AI in the field},'' in \emph{Proceedings of
  UKRAS21 Conference: Robotics at home}.\hskip 1em plus 0.5em minus 0.4em\relax
  UK-RAS, 2021.

\bibitem{zhivkov-et-al-machines:2023}
T.~Zhivkov, E.~I. Sklar, D.~Botting, and S.~Pearson, ``{5G on the Farm:
  Evaluating Wireless Network Capabilities and Needs for Agricultural
  Robotics},'' \emph{Machines}, vol.~11, no.~12, p. 1064, 2023.

\bibitem{cerf:tcp}
V.~Cerf and R.~Kahn, ``A protocol for packet network intercommunication,''
  \emph{IEEE Transactions on Communications}, vol.~22, no.~5, pp. 637--648,
  1974.

\bibitem{quigley:ros}
M.~Quigley, K.~Conley, B.~Gerkey, J.~Faust, T.~Foote, J.~Leibs, R.~Wheeler,
  A.~Y. Ng \emph{et~al.}, ``Ros: an open-source robot operating system,'' in
  \emph{ICRA workshop on open source software}, vol.~3.\hskip 1em plus 0.5em
  minus 0.4em\relax Kobe, Japan, 2009, p.~5.

\bibitem{beibei:ros-security}
B.~Yu, M.~Hu, Z.~Sun, and B.~Chen, ``Data tampering attack design for ros-based
  object detection and tracking robotic platform,'' in \emph{2021 International
  Conference on Control, Automation and Information Sciences (ICCAIS)}, 2021,
  pp. 159--164.

\bibitem{mqtt:oasis}
\BIBentryALTinterwordspacing
M.~Oasis Open~technical team, ``{MQTT Version 5.0},'' {Oasis Open}, Technical
  Report (TR), March 2019. [Online]. Available:
  \url{https://docs.oasis-open.org/mqtt/mqtt/v5.0/mqtt-v5.0.pdf}
\BIBentrySTDinterwordspacing

\bibitem{chen2021fogros}
K.~E. Chen, Y.~Liang, N.~Jha, J.~Ichnowski, M.~Danielczuk, J.~Gonzalez,
  J.~Kubiatowicz, and K.~Goldberg, ``{FogROS: An Adaptive Framework for
  Automating Fog Robotics Deployment},'' in \emph{2021 IEEE 17th International
  Conference on Automation Science and Engineering (CASE)}.\hskip 1em plus
  0.5em minus 0.4em\relax IEEE, 2021, pp. 2035--2042.

\bibitem{antevski2018enhancing}
K.~Antevski, M.~Groshev, L.~Cominardi, C.~J. Bernardos, A.~Mourad, and
  R.~Gazda, ``Enhancing edge robotics through the use of context information,''
  in \emph{Proceedings of the Workshop on Experimentation and Measurements in
  5G}, 2018, pp. 7--12.

\bibitem{hayat2021edge}
S.~Hayat, R.~Jung, H.~Hellwagner, C.~Bettstetter, D.~Emini, and D.~Schnieders,
  ``Edge computing in {5G} for drone navigation: What to offload?'' \emph{IEEE
  Robotics and Automation Letters}, vol.~6, no.~2, pp. 2571--2578, 2021.

\bibitem{huang2022edge}
P.~Huang, L.~Zeng, X.~Chen, K.~Luo, Z.~Zhou, and S.~Yu, ``Edge robotics:
  Edge-computing-accelerated multi-robot simultaneous localization and
  mapping,'' \emph{IEEE Internet of Things Journal}, 2022.

\bibitem{Zahidi2021}
U.~A. Zahidi and G.~Cielniak, ``Active learning for crop-weed discrimination by
  image classification from convolutional neural network's feature pyramid
  levels,'' in \emph{Computer Vision Systems}, M.~Vincze, T.~Patten, H.~I.
  Christensen, L.~Nalpantidis, and M.~Liu, Eds.\hskip 1em plus 0.5em minus
  0.4em\relax Cham: Springer International Publishing, 2021, pp. 245--257.

\bibitem{RedbrickAI2021}
RedbrickAI, ``Redbrickai ai annotation tools,'' \url{https://redbrickai.com/},
  2021.

\bibitem{eval_metrics_2021}
\BIBentryALTinterwordspacing
R.~Padilla, W.~L. Passos, T.~L.~B. Dias, S.~L. Netto, and E.~A.~B. da~Silva,
  ``A comparative analysis of object detection metrics with a companion
  open-source toolkit,'' \emph{Electronics}, vol.~10, no.~3, 2021. [Online].
  Available: \url{https://www.mdpi.com/2079-9292/10/3/279}
\BIBentrySTDinterwordspacing

\bibitem{augustyn2015real}
J.~Augustyn, ``Real-time performance of hybrid mobile robot control utilizing
  usb protocol,'' \emph{International Journal of Advanced Robotic Systems},
  vol.~12, no.~2, p.~14, 2015.

\bibitem{Energyusage2022}
Z.~Wang, ``Energyusage python package,''
  \url{https://pypi.org/project/energyusage/}, 2022.

\bibitem{Fastpick2022a}
M.~Hanheide, J.~Dichtl, U.~A. Zahidi, A.~Khan, T.~Zhivkov, D.~Li, G.~Cielniak,
  and A.~Ghalamzan, ``University of lincoln, fastpick project: Source code,''
  \url{https://github.com/LCAS/drydock_ros}, 2022.

\bibitem{Fastpick2022b}
M.~Hanheide, A.~Khan, U.~A. Zahidi, S.~Parsa, T.~Zhivkov, D.~Li, J.~Dichtl,
  G.~Cielniak, E.~I. Sklar, S.~Pearson, and A.~Ghalamzan, ``University of
  lincoln, fastpick project: semantic segmentation models and field-tests
  videos,'' \url{https://lncn.ac/fastpick}, 2022.

\end{thebibliography}

\section{Appendix}

\subsection{Normality test of network communication data}

We compute the Shapiro-Wilk statistic (W) on network communication data. If W is close to 1.0, then the data is likely normal. If $p < 0.01$, then we have $99$ \% confidence in the results (i.e. there is less than $1$\% probability that the results (W) occurred by chance). 

Table \ref{tab_5g_client_normality}--\ref{tab_wifi_server_normality} shows the normality test for MQTT and TCPROS protocols with different QoS configurations at client and server sides, respectively. The entries in the tables below that are in bold do not pass the Shapiro-Wilk test for normalcy.

\begin{table}[htpb!]
    \centering
    \begin{tabular}{rrrr}
                            \textbf{Communication type} &	  \textbf{n} &	     \textbf{W} &	         \textbf{p} \\ 
    \hline
        MQTT QoS0 rxkBs &	104 &	 0.983 &	\textbf{0.22021087} \\ 
        MQTT QoS0 txkBs &	104 &	 0.950 &	0.00061510 \\ 
        MQTT QoS1 rxkBs &	 93 &	 0.447 &	0.00000000 \\ 
        MQTT QoS1 txkBs &	 93 &	 0.587 &	0.00000000 \\ 
        TCPROS    rxkBs &	 99 &	 0.982 &	\textbf{0.21213713} \\ 
        TCPROS    txkBs &	 99 &	 0.980 &	\textbf{0.12610261} \\ 
    \hline
    \end{tabular}
    \caption{\textbf{5G communication at Client (Robot) side}}. 
    \label{tab_5g_client_normality}
\end{table}

\begin{table}[htpb!]
    \centering
\begin{tabular}{rrrr}
                            \textbf{Communication type} &	  \textbf{n} &	     \textbf{W} &	         \textbf{p} \\ 
\hline
 
  MQTT QoS0 rxkBs &	 99 &	 0.986 &	\textbf{0.37013268} \\ 
  MQTT QoS0 txkBs &	 99 &	 0.945 &	0.00039823 \\ 
  MQTT QoS1 rxkBs &	 99 &	 0.992 &	\textbf{0.82304007} \\ 
  MQTT QoS1 txkBs &	 99 &	 0.951 &	0.00100388 \\ 
  TCPROS rxkBs &	 99 &	 0.976 &	\textbf{0.06729713} \\ 
  TCPROS txkBs &	 99 &	 0.950 &	0.00093943 \\ 
\hline
\end{tabular}
    \caption{\textbf{WiFi communication at Client (Robot) side}}. 
    \label{tab_wifi_client_normality}
\end{table}

\begin{table}[htpb!]
    \centering
\begin{tabular}{rrrr}
                            \textbf{Communication type} &	  \textbf{n} &	     \textbf{W} &	         \textbf{p} \\ 
\hline
    MQTT QoS0 rxkBs &	 99 &	 0.908 &	0.00000363 \\ 
    MQTT QoS0 txkBs &	 99 &	 0.984 &	\textbf{0.25549772} \\ 
    MQTT QoS1 rxkBs &	 99 &	 0.975 &	\textbf{0.05627089} \\ 
    MQTT QoS1 txkBs &	 99 &	 0.988 &	\textbf{0.48127228} \\ 
    TCPROS rxkBs &	 99 &	 0.935 &   0.00010156 \\ 
    TCPROS txkBs &	 99 &	 0.970 &  \textbf{0.02148055}\\ 
\hline
\end{tabular}
    \caption{\textbf{5G communication at Server side}}. 
    \label{tab_5G_server_normality}
\end{table}

\begin{table}[htpb!]
    \centering
\begin{tabular}{rrrr}
                            \textbf{Communication type} &	  \textbf{n} &	     \textbf{W} &	         \textbf{p} \\ 
\hline
 
  MQTT QoS0 rxkBs &	 99 &	 0.931 &	0.00006434 \\ 
  MQTT QoS0 txkBs &	 99 &	 0.991 &	\textbf{0.75420296} \\ 
  MQTT QoS1 rxkBs &	 99 &	 0.573 &	0.00000000 \\ 
  MQTT QoS1 txkBs &	 99 &	 0.192 &	0.00000000 \\ 
  TCPROS rxkBs &	 99 &	 0.926 &	0.00003293 \\ 
  TCPROS txkBs &	 99 &	 0.976 &	\textbf{0.06144753} \\ 
\hline
\end{tabular}
    \caption{\textbf{WiFi communication at Server side}}. 
    \label{tab_wifi_server_normality}
\end{table}

\subsection{Operational parameters of 5G network}

The 5G configuration has limitations such as the current Time Division Duplexing (TDD) and carrier bandwidth, as shown in Table \ref{tab:5gsystem}, are fixed and the 5G frequency range is subject to UK OFCOM licensing\footnote{Link - \url{https://www.ofcom.org.uk/__data/assets/pdf_file/0016/103309/uk-fat-2017.pdf}}. Here, DL and UL stand for \textit{download} and \textit{upload} respectively, and are used typically to denote throughput speed or refer to modulation.

\begin{table}[!htpb]
\centering
\begin{tabular}{|cc|}
\hline
\multicolumn{1}{|c|}{\textbf{Specification}} & \textbf{Description}       \\ \hline
\multicolumn{1}{|c|}{5G Band N77}            & 3800MHz-4100MHz            \\ \hline
\multicolumn{1}{|c|}{Carrier Bandwidth}      & 100MHz                     \\ \hline
\multicolumn{1}{|c|}{Modulation}             & 256(DL)/64(UL) QAM         \\ \hline
\multicolumn{1}{|c|}{Transmit power}         & 5W per Tx path (4Tx paths) \\ \hline
\multicolumn{1}{|c|}{MIMO layers}            & 4x2 closed-loop MIMO \\ \hline
\multicolumn{1}{|c|}{TDD  (UL:DL) ratio}     & 3/7                        \\ \hline
\end{tabular}
\caption{\textbf{5G stand-alone N77 network configuration.}}
\label{tab:5gsystem}
\end{table}
\end{document}